\documentclass{article}

\usepackage[main, final] {neurips_2026}
\usepackage[utf8]{inputenc} 
\usepackage[T1]{fontenc}    
\usepackage{hyperref}       
\usepackage{url}            
\usepackage{booktabs}       
\usepackage{amsfonts}       
\usepackage{nicefrac}       
\usepackage{microtype}      
\usepackage{xcolor}         
\usepackage{multirow}
\usepackage{bbm}
\usepackage{amsmath} 
\usepackage{amssymb} 
\usepackage[table]{xcolor}
\usepackage{graphicx} 
\usepackage{wrapfig}
\usepackage{booktabs}
\usepackage{tcolorbox}

\newcommand{\tool}{{ToolSearcher}}
\definecolor{table-blue}{RGB}{220, 230, 242}

\title{ToolSearcher: Optimizing Tool Selection at Scale via Reinforcement Learning}

\author{%
  Zhenlong Dai\\
 Zhejiang University\\
\texttt{zhenlongdai@zju.edu.cn} \\
  \And
  Xujie Song \\
  Zhejiang University \\
  \texttt{songxujie@zju.edu.cn} \\
  \AND
  Zitong Wang \\
  Ant Group \\
  \texttt{yesi.wzt@antgroup.com} \\
  \And
  Tong Niu \\
  Ant Group \\
  \texttt{niutong.niu@antgroup.com} \\
  \And
  Jian Liu\thanks{Co-corresponding authors.} \\
  Ant Group \\
  \texttt{rex.lj@antgroup.com} \\
    \And
  Weiqiang Wang \\
  Ant Group \\
  \texttt{weiqiang.wwq@antgroup.com} \\
  \And
  Xiu Tang \\
  Zhejiang University \\
  \texttt{tangxiu@zju.edu.cn} \\
  \And
  Sai Wu \\
  Zhejiang University  \\
  \texttt{wusai@zju.edu.cn} \\
  \And
  Chang Yao\footnotemark[1]  \\
  Zhejiang University  \\
  \texttt{changy@zju.edu.cn} \\
    \And
  Jingyuan Chen \\
  Zhejiang University  \\
  \texttt{jingyuanchen@zju.edu.cn} 
}

\begin{document}

\maketitle

\begin{abstract}
Large language models (LLMs) excel at natural language processing but struggle to interact with external environments. 
Tool learning provides a promising way to extend LLMs into actionable agents, where tool selection is a critical prerequisite for successful tool use. 
Existing work often assumes a small or predefined set of tools, leaving large-scale tool selection underexplored. 
Real-world repositories contain a vast and diverse array of tools, making it difficult for LLMs to effectively search, distinguish, and compose tools under context-length constraints.
We identify large-scale tool selection as a new challenge for agentic reinforcement learning, highlighting that existing RL methods for knowledge-based question answering are inadequate for selecting tools while considering compatibility.
To address this challenge, we propose \tool, a novel RL framework for effective multi-turn search and fine-grained optimization in large-scale tool selection. 
Specifically, we introduce category-constrained tool discrimination to improve the model's ability to distinguish functionally similar tools, event-level search modeling to explicitly optimize the discovery of target tools during multi-turn search, and trajectory-aligned credit allocation to provide fine-grained reward signals for different stages of the search-selection process. 
Extensive experiments on large-scale tool selection benchmarks demonstrate that \tool~consistently outperforms a set of strong baselines in challenging settings involving iterative search and complex tool composition. 
The code and the dataset are available at \url{https://github.com/zhenlongDai/ToolSearcher}.
\end{abstract}

\section{Introduction}
\label{Introduction}
Large language models (LLMs) have recently demonstrated remarkable capabilities in natural language processing (NLP) tasks~\cite{kumar2025llm,dai2024mpcoder,dai2025less}. 
Despite their impressive performance, pretrained LLMs often struggle to actively perceive and interact with external environments~\cite{dai2026enhancing,dai2026learner,liu2026egobench}, as their abilities are intrinsically constrained by NL interaction.
Given the essential role of tools in enabling LLMs to interact with external environments and to act as user agents for solving real-world tasks, tool learning has attracted increasing attention~\cite{wang2023mint}.
The typical process of tool learning comprises three key stages: tool selection, tool calling, and response generation~\cite{qu2025tool}. Existing studies mainly focus on tool calling (e.g., specifying function parameters, tool execution), while overlooking a crucial yet underexplored issue of tool selection: \textit{how to effectively search and select appropriate candidate tools from a large-scale and diverse tool repository according to specific task requirements.}

Recent studies~\cite{patil2025bfcl,wang2024llms,shi2025tool} typically assume a predefined or small set of candidate tools, which creates a substantial gap between these settings and real-world scenarios. 
Statistics from mainstream software and model ecosystems (e.g., PyPI, npm, and HuggingFace) show that both the number and functional diversity of available tools are already massive~\cite{zhou2024large,hu2025understanding}.
Due to context-length limitations, it is impractical to provide LLMs with documentation for all tools, making the search a necessary step for effectively using existing tool resources.
Moreover, large-scale tool repositories further exacerbate both functional and semantic similarities among tools. 
On one hand, many tools have similar semantics but differ in functionality. 
On the other hand, some tools offer similar functionality but have different interface designs (\textit{i.e.}, input constraints and output information).
Under this background, tool selection is inherently a complex process involving planning, search, and decision-making, requiring the coordination of multiple capabilities.

LLMs remain limited in autonomously organizing the capabilities required for tool selection.
Current research~\cite{schick2023toolformer,trivedi2023interleaving} utilizes search engines as part of the reasoning process, allowing LLMs to conduct multi-turn searches. 
Given the difficulty of collecting high-quality annotated trajectories and improving the model's adaptability to dynamic search environments~\cite{chu2025sft,asai2024self}, they optimize via agentic reinforcement learning (RL)~\cite{jin2025search,song2025r1}.
However, these methods are designed for knowledge-intensive questions that rely on iterative information supplementation, making them unsuitable for tool selection tasks that require reasoning about tool composability.
For tool selection, each search step needs to plan around a hypothesized tool chain and consider the composition of tools, then adjust the plan based on the search results.
Based on the above analysis, applying RL to tool selection in large-scale scenarios mainly faces two key challenges:
(1) \textbf{How to effectively perform multi-turn search over a large-scale tool repository}. 
Tool selection requires the model to search relevant tools, distinguish between functionally or semantically similar candidates, and reason about interface compatibility over multiple search steps.
However, existing RL-based methods do not explicitly model the tool search process.
(2) \textbf{Tool selection involves multiple capabilities, making joint learning and reward design difficult}.
RL methods typically rely on trajectory-level rewards that assign the same value to a trajectory, leading to sparse supervision and making it difficult to optimize fine-grained behaviors throughout the multi-turn process.


To tackle the above challenges, we propose a novel RL approach named \tool, which is designed to enable effective multi-turn search and fine-grained optimization for large-scale tool selection. 
To differentiate between functionally similar tools, we propose category-constrained tool discrimination, which creates a highly constrained challenge environment designed to enhance LLMs’ ability to understand and distinguish tool functionalities.
To explicitly model the tool search process, we propose event-level search modeling, which optimizes semantic search by concentrating on events that discover previously unsearched target tools, thereby enhancing the LLM's ability to plan and search for compatible tool compositions.
To facilitate the joint learning of multiple capabilities in tool selection, we design trajectory-aligned credit allocation, a fine-grained reward mechanism that assigns objective and quantifiable credit based on each sample’s progression within the search-selection trajectory. 
Unlike outcome-only rewards, this method evaluates samples at different stages, providing matched feedback that accurately reflects their progress for tool selection.

Our contributions are summarized as follows:
\begin{itemize}
    \item We identify large-scale tool selection as a novel challenge for agentic RL and demonstrate that existing RL methods designed for knowledge-oriented search are unsuited to multi-turn tool search and reasoning about tool composability.
    \item We propose \tool, a novel RL framework that explicitly models the multi-turn search process over large-scale tool repositories and provides fine-grained optimization signals for the diverse capabilities involved in tool selection.
    \item We conduct extensive experiments on large-scale tool selection benchmarks, where \tool~consistently outperforms other baselines in challenging scenarios requiring iterative search and complex tool composition.
\end{itemize}

\section{Related Work}

\subsection{Retrieval-Augmented Generation}
Retrieval-Augmented Generation (RAG) enhances language models by grounding their responses in external knowledge sources~\cite{karpukhin2020dense}. 
RAG typically follows a retrieval-then-generation pipeline~\cite{gao2023retrieval}, where a search engine fetches relevant information based on the query, which is then combined with the query and fed into the LLM.
Early RAG approaches~\cite{kwiatkowski2019natural,joshi2017triviaqa} focused on retrieval tasks in which answers reside within individual documents. 
Recent advances~\cite{asai2023self,jiang2023active,trivedi2022musique} have extended RAG to multi-hop reasoning, where answers reside across multiple documents.
Although these methods improve reasoning over small document sets, they remain constrained by top-k retrieval paradigms, which fundamentally limit their ability to achieve complete document coverage for a large corpus~\cite{luo2025hypergraphrag}. 
These paradigms can lead to challenges such as retrieving irrelevant information~\cite{jin2024long} and failing to provide sufficiently relevant context~\cite{jiang2023active}. 
Tool-integrated agents~\cite{trivedi2023interleaving,schick2023toolformer} have expanded the possibilities of RAG by treating the search engine as a tool, enabling LLMs to interleave reasoning and action with external tools~\cite{li2024llatrieval}. 
However, these training-free agentic approaches show promise but are limited by the quality of prompting strategies and lack the ability to learn from task-specific feedback.

\subsection{Agentic Reinforcement Learning}
Agentic reinforcement learning~\cite{zhang2025landscape} enables agents to make sequential decisions aimed at maximizing cumulative rewards. 
In LLM tuning, RL has become a highly effective paradigm through RLHF~\cite{kaufmann2023survey}, implemented with PPO~\cite{schulman2017proximal}.
To address the complexity inherent in PPO, methods like DPO~\cite{rafailov2023direct} and SimPO~\cite{meng2024simpo} have been proposed as more efficient alternatives.
While these methods offer computational efficiency, they suffer from off-policy issues~\cite{pang2024iterative} and do not consistently match the performance of pure RL approaches~\cite{jin2025search}.
An effective RL method is Group Relative Policy Optimization (GRPO)~\cite{shao2024deepseekmath}, which significantly improves LLMs' reasoning by estimating baselines from group scores.
Building on this, GSPO~\cite{zheng2025group} enables stable optimization in multi-turn RL by handling precision discrepancies with sequence-level likelihoods.
GDPO~\cite{liu2026gdpo} decouples reward normalization to preserve relative differences to enhance multi-reward optimization.
More recent work~\cite{jin2025search,li2025search} has applied RL to multi-step retrieval, but they primarily optimize local retrieval over small document subsets~\cite{edge2024local,luo2025hypergraphrag}. 
MARAG-R1~\cite{luo2025marag} expands in this direction by leveraging RLOO~\cite{ahmadian2024back} to achieve global retrieval tasks that require comprehensive document coverage.
Despite these advances, RL for tool selection in large-scale repositories remains largely unexplored, particularly in distinguishing between similar tools and learning compatible tool combinations.

\section{Method}
\label{Method}
LLMs remain limited in autonomously organizing the capabilities required for tool selection. 
To address this, we propose an RL-based approach, \tool~, that enhances tool selection within large-scale tool repositories. 
In this section, we first define the tool selection task (Section~\ref{sec:task_define}) and provide the preliminary information (Section~\ref{sec:Preliminary}). 
We then propose category-constrained tool discrimination (Section~\ref{sec:Category}) to improve the differentiation of functionally similar tools in large-scale tool repositories.
Moreover, We introduce our RL approach for tool selection (Section~\ref{sec:RL}), which incorporates event-level search modeling (Section~\ref{sec:Event})  and trajectory-aligned credit allocation (Section~\ref{sec:Aligned})  to effectively guide the learning process in searching and identifying compatible tool combinations that meet task requirements.

\subsection{Task Definition}
\label{sec:task_define}
In a large-scale tool repository \(\mathcal{T}\), given a user requirement $q$ and a search engine $ \mathcal{R}$, the goal is to identify a tool set $\mathcal{T}_s$ that satisfies task requirements. 
For each tool $T \in \mathcal{T}$, The document $d_{T} = (f, \mathcal{I}, \mathcal{O}_t) $ contains a function description $f$ and an interface specification, including its input constraints  $\mathcal{I}$ and output schema $\mathcal{O}$. 
Formally, the objective of tool selection is to identify a set of tools $\mathcal{T}_s = \{T_1,T_2, \dots, T_n\}$ by a model $\pi_{\theta}$ such that the conditional probability $P_{\theta}(\mathcal{T}_s | q,\mathcal{T}, \mathcal{R})$ is maximized.

\subsection{Preliminary}
\label{sec:Preliminary}
To enable the LLM to use the search engine tool, the schema of the search engine $\mathcal{R}$ is included in its prompt.
The schema defines the search engine's callable interface, including the tool identifier and the required query argument fields,  enabling the LLM to generate well-formed tool invocations in a structured format.
At each step, the LLM produces a tool call enclosed by the special tokens \texttt{<tool\_call>} and \texttt{</tool\_call>}, where the enclosed content is a JSON object whose fields conform to the predefined schema.
The search engine then executes the corresponding query and returns the retrieved results to the LLM as external feedback.
Based on the feedback, the LLM can proceed to the next search round by following the same process.

\subsection{Category-Constrained Tool Discrimination}
\label{sec:Category}
Large-scale tool repositories exacerbate the similarity among tools, resulting in search outcomes that often include tools with similar functionality or semantics. 
This requires the model to understand and distinguish between similar tools within documents at each search step. 
To address this challenge, we propose category-constrained tool discrimination, which leverages tool categories to construct a strongly constrained challenge environment for tool search. 
The core idea is to transform tool retrieval from a global search over the entire repository into a category-constrained search within a functionally similar subspace.

\begin{figure*}[t]
\centering
\includegraphics[width=\textwidth]{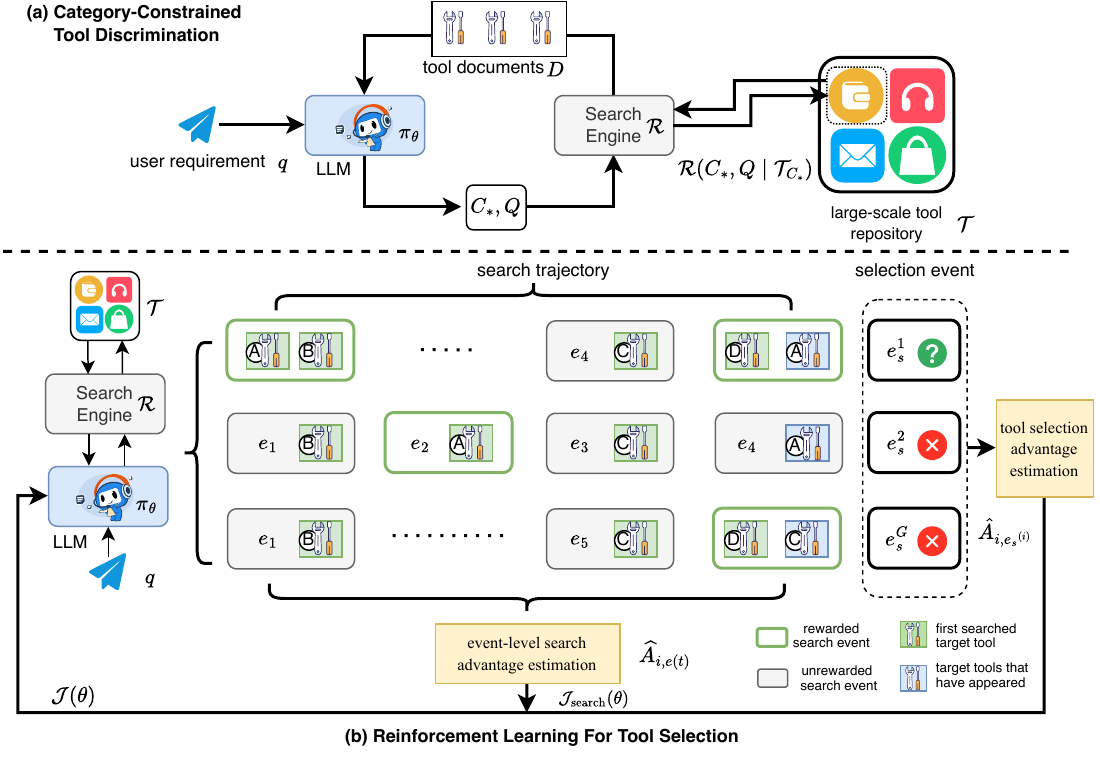}
\vspace{-2.0em}
\caption{Overview of \tool. (a) Illustration of the category-constrained tool discrimination. (b) Illustration of RL for tool selection that contains event-level search modeling and trajectory-aligned credit allocation.
} 
\label{fig:model}
\end{figure*}

Specifically, as shown in Fig.~\ref{fig:model}, each tool \(T \in \mathcal{T}\) is assigned to a specific category \(C\) based on its functionality (\textit{e.g.}, social, sports, finance). 
\(\mathcal{T}\) is represented as \(\{\mathcal{T}_{C_1}, \mathcal{T}_{C_2}, \dots, \mathcal{T}_{C_m}\}\), where each subset \(\mathcal{T}_{C_i}\) corresponds to a category-specific set of tools. 
We equip the search engine \(\mathcal{R}\) with a subcategory search function, which is integrated into its schema as shown in Appendix~\ref{Prompt}. 
At each search, 
Each search step can be formalized as follows:
\begin{equation}
D =  \mathcal{R}(C_*, Q \mid \mathcal{T}_{C_*}),
\end{equation}
where $D$ is the tool documents that are searched within a specific subset of tools \(\mathcal{T}_{C_*}\) based on a given category \(C_*\) and query $Q$. 
Accessing tools by category in large-scale repositories creates a challenging scenario, which greatly increases the similarity between search tools. This forces LLMs to focus on distinguishing functionally similar tools, rather than optimizing the query for fine-grained semantic search.
We employ the RL training approach for tool selection described in Section~\ref{sec:RL}, applying it in a two-phase training process.
In the first phase, the LLM is trained using 30\% of the data in a subcategory-specific search setting, focusing on distinguishing tools with functional similarities. 
In the second phase, the LLM is trained in a global search setting, learning to perform fine-grained semantic searches across the entire tool repository. 
Based on the progressive learning strategy, we guide LLMs to discriminate among similar tools and search effectively within large-scale tool repositories.

\subsection{Reinforcement Learning For Tool Selection}
\label{sec:RL}

\subsubsection{Event-level Search Modeling.}
\label{sec:Event}

Tools often exhibit dependencies and interface constraints, leading to non-executable tool chains when tool compositions are incompatible. 
To reliably identify compatible tool combinations, search and planning operate collaboratively during the multi-turn search process:
The LLM needs to adjust its plan for a candidate tool chain by reasoning about tool composition based on current search results.
However, existing RL-based approaches do not explicitly model the tool search process, limiting the ability to improve the LLM's search for compatible tool combinations.
To address this challenge, we propose event-level search modeling, which integrates the search process with the RL objective by measuring the significance of individual search events for tool composition within each trajectory, thereby enhancing LLMs' ability to search and plan for compatible tool combinations.


Specifically, as shown in Fig.~\ref{fig:model}, we explicitly model the search trajectory $\mathcal{S}$ as a sequence of search events $\mathcal{E}= (e_1, e_2, \dots, e_k)$,
where $e_k$ denotes the event at step $k$.
Each event is defined as $e_k = (a_k, \mathcal{R}(a_k))$, where $a_k$ is the model action at step $k$, which is a tool call to the search engine, and $\mathcal{R}(a_k)$ denotes the search tool sets. 
To prevent LLMs from learning and memorizing tokens from retrieved tool documents, we follow prior work~\cite{jin2025search} and apply loss masking to retrieved tokens, ensuring that the policy-gradient objective is computed only over LLM-generated tokens and excludes retrieved content $\mathcal{R}(a_k)$ from the optimization process.
The set of events $\mathcal{E}^*$ that successfully search previously uncovered target tools for the first time, and the target tools $\mathcal{T}_{e_j}$ searched by specific events $e_j$, are defined as:
\begin{equation}
\footnotesize
\begin{split}
\mathcal{E}^* &= \left\{ e_j \;\middle|\; 1 \le j \le k, \;  \mathcal{T}_{e_j} \neq \emptyset \right\}, \\
\mathcal{T}_{e_j} &= \mathcal{M}(e_j) \cap \left( \mathcal{T}_s \setminus \mathcal{M}(\mathcal{E}_{<j}) \right),
\end{split}
\end{equation}
where \(\mathcal{M}(\cdot)\) denotes the search tools obtained from the search events.
Considering the potential inaccuracies caused by the search engine's performance and the necessity of enhancing the model's ability to recognize target tools, we optimize only search events $\mathcal{E}^*$ that involve previously unsearched target tools and avoid optimizing events that re-search tools already searched.
To enhance LLMs’ reasoning about tool interface compatibility within the current search process, we maximize the following event-level optimization objective:
\begin{equation}
\footnotesize
\begin{split}
\mathcal{J}_{\text{search}}(\theta)=\mathbb{E}_{\mathcal{S}_{\mathrm{search}} \sim \pi_\theta}
\left[
\prod_{j=1}^{k} P_{\theta}(e_j | \mathcal{E}_{<j})^{\mathbbm{1}_{e_j \in \mathcal{E}^*}}
\right].
\end{split}
\end{equation}
\label{eq1}
Inspired by group relative policy optimization (GRPO)~\cite{shao2024deepseekmath}, we achieve the above objective by performing event-level advantage optimization across search events.
Specifically, for each user requirement $q$ that samples drawn from the dataset $\mathcal{D}$, the policy LLM $\pi_{\theta_{\text{old}}}$ generate $G$ search trajectories $\{\mathcal{S}_i\}_{i=1}^G$, the RL objective utilizing search engine $\mathcal{R}$ is formulated  as follows:
\begin{equation}
\footnotesize
\begin{split}
    &\mathcal{J}_{\text{search}}(\theta) = \mathbb{E}{[q\sim \mathcal{D}, \{\mathcal{S}_i\}_{i=1}^G \sim \pi_{\theta_{\text{old}}}(\mathcal{S}|q,\mathcal{T}, \mathcal{R})]}  \\
    & \frac{1}{G}\sum_{i=1}^G\frac{1}{|\mathcal{S}_i|} \sum_{t=1}^{|\mathcal{S}_i|} \left\{ \min \left[ w_{i,t}(\theta,\mathcal{R}) \hat{A}_{i,e(t)}, \text{clip} \left( w_{i,t}(\theta,\mathcal{R}), 1 - \epsilon, 1 + \epsilon \right)  \hat{A}_{i,e(t)} \right] - \beta \mathbb{D}_{\text{KL}}\left(\pi_{\theta} || \pi_{\text{ref}}\right)\right\} ,
\end{split}
\label{eq:GRPO-obj}
\end{equation}
where $\epsilon$ and $\beta$ are hyper-parameters, $\mathbb{D}_{\text{KL}}$ is a KL penalty term.
The importance ratio $w_{i,t}(\theta,\mathcal{R})$ and event-level search advantage $\widehat{A}_{i,e(t)}$ of token $\mathcal{S}_{i,t}$ are:
\begin{align}
\footnotesize
\begin{split}
&w_{i,t}(\theta,\mathcal{R})=
\frac{\pi_{\theta}(\mathcal{S}_{i,t}\mid q,\mathcal{R},\mathcal{S}_{i,<t})}
{\pi_{\theta_{\text{old}}}(\mathcal{S}_{i,t}\mid q,\mathcal{R},\mathcal{S}_{i,<t})},\\
&\widehat{A}_{i,e(t)}=
\begin{cases}
\displaystyle
\max_{T \in \mathcal{T}_{e(t)}}
\left[
\frac{
r(T,\mathcal{S}_i)-\mathrm{mean}\left(\{r(T,\mathcal{S}_j)\}_{j=1}^{G}\right)
}{
\mathrm{std}\left(\{r(T,\mathcal{S}_j)\}_{j=1}^{G}\right)
}\right],
& e(t)\in \mathcal{E}_i^*, \\
0, & \text{otherwise}.
\end{cases}
\end{split}
\end{align}
where $e(t)$ denotes the search event where the token $\mathcal{S}_{i,t}$ is located,
$r$ is a binary function that indicates whether the target tool $T$ has been searched in the search trajectory $\mathcal{S}$ (1 if searched, 0 otherwise).
The event-level search advantage $\widehat{A}_{i,e(t)}$ is defined as the maximum advantage within the set of target tools $\mathcal{T}_{e(t)}$, which are first searched by the event $e(t)$ in the trajectory $\mathcal{S}_i$. 
By comparing the advantage gap between a search event and other trajectories, the event-level search advantage measures the significance of each search event within its respective trajectory.
Therefore, we explicitly model the impact of search events on the RL objective by adjusting the gradient of each event within the trajectory (see Appendix~\ref{Gradient} for gradient analysis).

\subsubsection{Trajectory-Aligned Credit Allocation.}
\label{sec:Aligned}
Tool selection is a complex process involving planning, searching, and decision-making. 
Even among trajectories for the same task, progress may differ substantially: some complete the task, others gather sufficient tool information yet still struggle with tool selection, while some remain at an earlier stage of the search process. 
Therefore, a desirable credit assignment should account for the stage of each trajectory and allocate rewards that drive progress toward successful tool selection.
To address this challenge, we propose trajectory-aligned credit allocation, which automatically focuses learning signals on unmastered steps by removing rewards for selection events associated with unfinished tool searches and rewards for search events that are mastered in trajectories.

Specifically, for trajectories $\{\mathcal{S}_i, e_{\text{s}}^{(i)}\}_{i=1}^G$, the final RL objective of $i$-th trajectory  is formulated  as follows:
\begin{equation}
\footnotesize
\begin{split}
     &\mathcal{J}^{(i)}(\theta)= \mathcal{J}^{(i)}_{\text{search}}(\theta) +\mathbb{E}{[e_{\text{s}}^{(i)} \sim \pi_{\theta_{\text{old}}}(e_{\text{s}}^{(i)} | q,\mathcal{S}_i)]} \\
     &\sum_{t=1}^{|e_{\text{s}}^{(i)}|} \min 
     [ \frac{\pi_{\theta}(e_{\text{s}, t}^{(i)}\mid q,\mathcal{S}_{i},e_{\text{s}, <t}^{(i)})}
{\pi_{\theta_{\text{old}}}(e_{\text{s}, t}^{(i)}\mid q,\mathcal{S}_{i},e_{\text{s}, <t}^{(i)})} \hat{A}_{i,e_{\text{s}}^{(i)}}, \text{clip} ( \frac{\pi_{\theta}(e_{\text{s}, t}^{(i)}\mid q,\mathcal{S}_{i},e_{\text{s}, <t}^{(i)})}
{\pi_{\theta_{\text{old}}}(e_{\text{s}, t}^{(i)}\mid q,\mathcal{S}_{i},e_{\text{s}, <t}^{(i)})}, 1 - \epsilon, 1 + \epsilon )  \hat{A}_{i,e_{\text{s}}^{(i)}} ],
\end{split}
\label{eq:GRPO-both}
\end{equation}
where the tool selection advantage \(\hat{A}_{i,e_{\text{s}}^{(i)}}\) is computed by the selection reward $r_s$ if the search process \(\mathcal{S}_i\) includes all the target tools \(\mathcal{T}_s\):
\begin{align}
\footnotesize
\begin{split}
\hat{A}_{i,e_{\text{s}}^{(i)}} =
\begin{cases}
\displaystyle
\frac{
r_s(\mathcal{T}_s, e_{\text{s}}^{(i)}|\mathcal{S}_i)-\mathrm{mean}(\{r_s(\mathcal{T}_s, e_{\text{s}}^{(i)}|\mathcal{S}_i)\}_{j=1}^{G})
}{
\mathrm{std}\left(\{r_s(\mathcal{T}_s, e_{\text{s}}^{(i)}|\mathcal{S}_i)\}_{j=1}^{G}\right)
},
& \mathcal{M}(\mathcal{S}_i)  \cap \mathcal{T}_s = \mathcal{T}_s, \\
0, & \text{otherwise}.
\end{cases}
\end{split}
\end{align}
The selection reward $r_s$ is a binary function that equals 1 if the selection event \(e_{\text{s}}^{(i)}\) matches the target tools \(\mathcal{T}_s\), and 0 otherwise.
For the selection stage, this RL objective prevents the model from making unrealistic guesses for tool selections based on past training outcomes or tools that were searched during the search trajectory.
For the search stage, the event-level search advantage $\widehat{A}_{i,e(t)}$ is set to zero for search events involving corresponding target tools that have already been mastered within the group.
As a result, as shown in Fig.~\ref{fig:model}, $\mathcal{J}^{(i)}_{\text{search}}(\theta)$ eliminates the reward for search capabilities related to tools that are mastered within the group (\textit{i.e.}, colored in gray), 
while rewarding search events within trajectories that outperform other trajectories in searching for the target tools (\textit{i.e.},  colored in green).
Based on the above design, we allocate rewards for multiple search events and selection events in trajectories at different stages of progress, ensuring that the rewards correspond to the advancement made relative to overall progress.

\section{Experimental Setup}
\label{Experimental_Setup}
\subsection{Datasets and Metrics.}
To evaluate the effectiveness of \tool~in tool selection at scale, we use StableToolBench~\cite{guo2024stabletoolbench, qin2023toolllm} as the training and test dataset. StableToolBench contains the APIs of 16k tools across 49 diverse categories and covers various scenarios: \textbf{single-tool instructions (I1)}, \textbf{intra-category multi-tool instructions (I2)}, and \textbf{intra-collection multi-tool instructions (I3)}.
To evaluate the robustness on out-of-distribution datasets, we conduct the evaluation on AppWorld~\cite{trivedi2024appworld}, which is a high-quality execution environment featuring 9 day-to-day apps operable via 457 APIs and populated with realistic digital activities simulating the lives of approximately 100 fictitious users.
To evaluate tool selection, we assess the results using several metrics: F1, Recall, Precision, and Match, which measure the accuracy of the selected tools compared to the ground truth. 
Additionally, we use SRecall to measure the effectiveness of retrieving the target tools during the search stage.
For the AppWorld evaluation, we follow their predefined method and utilize FullCodeRefl with gpt-5-mini~\footnote{https://developers.openai.com/api/docs/models/all} as the agent to generate code and execute tasks based on the tool selection results.
The key metrics reported include Task Goal Completion (TGC) and Scenario Goal Completion (SGC). 
For further details on the datasets and metrics, please refer to Appendix~\ref{Dataset_Metric}.

\subsection{Baselines.}
We choose baselines from two paradigms of using search engines. 
For the retrieval-augmented generation paradigm, we adopt \textbf{RAG}~\cite{lewis2020retrieval} and \textbf{RAG\textsubscript{SFT}}, where the latter is the supervised fine-tuned (SFT)~\cite{chung2024scaling} model.
We further include \textbf{ToolEmbed}~\cite{lu2026tools}, which combines a fine-tuned Qwen3-Embedding-0.6B retriever with our supervised fine-tuned LLM.
For the paradigm that treats the search engine as an external tool, we denote the untrained baseline as \textbf{Multi-turns}. 
We also include several mainstream RL methods for comparison, including \textbf{Search-R1}~\cite{jin2025search}, \textbf{GSPO}~\cite{zheng2025group}, \textbf{GDPO}~\cite{liu2026gdpo}, \textbf{MARAG-R1}~\cite{luo2025marag}.
We additionally compare with two methods for tool-using agents: \textbf{ToolPO}~\cite{li2026deepagent}, the RL-based training method in DeepAgent for end-to-end interactive tool use, and \textbf{ToolQP}~\cite{fang2026beyond}, which relies on teacher-annotated plans to guide multi-step tool retrieval.
To enable autonomous multi-turn tool calls with the search engine, we choose Qwen2.5-7B-Instruct~\cite{qwen2} as the baseline, as it has strong and stable tool-calling capabilities for interacting with the search engine.
To further verify the feasibility of our method on smaller-scale models, we also adopt Qwen3-4B-Instruct~\cite{yang2025qwen3} as an additional baseline.
We also compare with a range of leading general-purpose LLMs to assess their tool-selection capabilities.
Further details are provided in Appendix~\ref{Baselines}.

\subsection{Implementation Details}
Considering the instruction-following and tool-calling capabilities, we conduct experiments on Qwen2.5-7B-Instruct~\cite{qwen2} and Qwen3-4B-Instruct~\cite{yang2025qwen3}.
For tool search, we use the tools from StableToolBench as the large-scale tool repository. 
Considering the limitation of the training context length, we set the number of retrieved tool documents to 5 per turn for all multi-turn methods. Based on the tool retrieval performance comparison, we finally adopt Qwen3-Embedding-0.6B~\cite{zhang2025qwen3} as the retriever for the search engine.
During training, we set the maximum number of interaction turns to 8 and sample 5 rollouts for each prompt. Further implementation details are provided in Appendix~\ref{Implementation}.

\begin{table}[t!]
\caption{Overall performance and per-scenario F1 scores on Stabletoolbench.}
\vspace{-0.5em}
\label{main_table}
\centering
\resizebox{0.90\textwidth}{!}{
\begin{tabular}{lcccccccccc}
\toprule
\multirow{2}{*}{Method} 
  & \multicolumn{4}{c}{Overall} 
  & \multicolumn{3}{c}{I1} 
  & \multicolumn{2}{c}{I2} 
  & \multicolumn{1}{c}{I3} \\
\cmidrule(lr){2-5} \cmidrule(lr){6-8} \cmidrule(lr){9-10} \cmidrule(lr){11-11}
  & F1 & Recall & Precision & Match 
  & Inst. & Tool &Cate.
  & Inst. & Cate.
  & Inst. \\
\midrule
\rowcolor{gray!10}
Gpt-5.5 & \textbf{0.569} & \textbf{0.572} & \textbf{0.590} & 0.336 & \textbf{0.675} & \textbf{0.635} & \textbf{0.645} & 0.474 & 0.454 & 0.316 \\
Claude-Opus-4-7 & \underline{0.560} & 0.556 & \underline{0.581} & \underline{0.369} & \underline{0.645} & \underline{0.580} & 0.597 & \textbf{0.550} & \textbf{0.460} & \textbf{0.410} \\
\rowcolor{gray!10}
Claude-Opus-4-6 & 0.552 & 0.555 & 0.565 & \textbf{0.388} & 0.623 & 0.549 & \underline{0.637} & \underline{0.517} & \underline{0.457} & \underline{0.408} \\
Qwen3.5-397B-A17B & 0.501 & \underline{0.566} & 0.491 & 0.259 & 0.576 & 0.554 & 0.555 & 0.470 & 0.400 & 0.288 \\
\rowcolor{gray!10}
Qwen3-235B-A22B-Instruct & 0.497 & 0.527 & 0.500 & 0.260 & 0.581 & 0.560 & 0.539 & 0.489 & 0.389 & 0.234 \\
Gpt-4.1-mini & 0.479 & 0.510 & 0.483 & 0.234 & 0.546 & 0.520 & 0.512 & 0.465 & 0.408 & 0.285 \\
\rowcolor{gray!10}
Qwen3.6-27B & 0.461 & 0.490 & 0.466 & 0.247 & 0.537 & 0.514 & 0.529 & 0.418 & 0.374 & 0.198 \\
Qwen3-14B & 0.443 & 0.492 & 0.430 & 0.182 & 0.537 & 0.446 & 0.485 & 0.407 & 0.388 & 0.256 \\
\rowcolor{gray!10}
Qwen3-32B & 0.423 & 0.475 & 0.410 & 0.178 & 0.496 & 0.459 & 0.476 & 0.374 & 0.360 & 0.213 \\
Qwen3-8B & 0.378 & 0.481 & 0.343 & 0.122 & 0.468 & 0.436 & 0.439 & 0.330 & 0.233 & 0.213 \\
\midrule
\rowcolor{gray!10}
\multicolumn{11}{c}{Qwen2.5-7B-Instruct}\\
RAG & 0.194 & 0.285 & 0.174 & 0.075 & 0.257 & 0.226 & 0.230 & 0.139 & 0.105 & 0.126 \\
\rowcolor{gray!10}
RAG\textsubscript{SFT} & 0.441 & 0.433 & 0.470 & 0.250 & 0.536 & 0.546 & 0.522 & 0.316 & 0.322 & 0.170 \\
ToolEmbed & 0.337 & 0.348 & 0.347 & 0.177 & 0.376 & 0.295 & 0.408 & 0.340 & 0.338 & 0.152 \\
\rowcolor{gray!10}
Multi-turns & 0.098 & 0.110 & 0.100 & 0.046 & 0.119 & 0.161 & 0.099 & 0.039 & 0.065 & 0.054 \\
Search-R1 & 0.327 & 0.311 & 0.361 & 0.152 & 0.415 & 0.321 & 0.320 & 0.321 & 0.297 & 0.194 \\
\rowcolor{gray!10}
GSPO & 0.444 & 0.417 & 0.505 & 0.221 & 0.535 & 0.512 & \underline{0.565} & 0.331 & 0.278 & 0.196 \\
GDPO & \underline{0.496} & 0.487 & \underline{0.524} & \underline{0.255} & \underline{0.577} & \textbf{0.567} & 0.555 & 0.427 & \underline{0.404} & 0.228 \\
\rowcolor{gray!10}
MARAG-R1 & 0.448 & 0.425 & 0.500 & 0.204 & 0.567 & 0.537 & 0.557 & 0.305 & 0.279 & 0.211 \\
ToolPO & 0.391 & 0.420 & 0.388 & 0.191 & 0.461 & 0.405 & 0.382 & \underline{0.452} & 0.346 & 0.177 \\
\rowcolor{gray!10}
ToolQP & 0.440 & \underline{0.493} & 0.432 & 0.222 & 0.503 & 0.450 & 0.504 & 0.433 & 0.362 & \underline{0.260} \\
\rowcolor{table-blue}
\tool & \textbf{0.513} & \textbf{0.511} & \textbf{0.534} & \textbf{0.278} & \textbf{0.595} & \underline{0.561} & \textbf{0.577} & \textbf{0.485} & \textbf{0.405} & \textbf{0.294} \\
\midrule
\rowcolor{gray!10}
\multicolumn{11}{c}{Qwen3-4B-Instruct}\\
RAG & 0.387 & \textbf{0.540} & 0.384 & 0.220 & 0.438 & 0.426 & 0.432 & 0.324 & 0.356 & 0.221 \\
\rowcolor{gray!10}
RAG\textsubscript{SFT} & 0.427 & 0.416 & 0.463 & 0.235 & 0.558 & 0.460 & 0.447 & 0.391 & 0.352 & 0.167 \\
ToolEmbed & 0.449 & 0.447 & 0.472 & 0.284 & \textbf{0.637} & \textbf{0.625} & \textbf{0.625} & 0.413 & 0.364 & 0.271 \\
\rowcolor{gray!10}
Multi-turns & 0.408 & 0.439 & 0.410 & 0.169 & 0.454 & 0.506 & 0.420 & 0.317 & 0.349 & 0.272 \\
Search-R1 & 0.505 & 0.504 & 0.516 & \underline{0.307} & 0.583 & 0.604 & 0.541 & 0.409 & 0.465 & 0.215 \\
\rowcolor{gray!10}
GSPO & 0.473 & 0.453 & 0.518 & 0.258 & 0.576 & 0.572 & 0.576 & 0.305 & 0.362 & 0.221 \\
GDPO & \underline{0.518} & 0.513 & \underline{0.537} & 0.305 & 0.576 & 0.616 & 0.574 & 0.393 & \textbf{0.505} & 0.238 \\
\rowcolor{gray!10}
MARAG-R1 & 0.465 & 0.439 & 0.527 & 0.225 & 0.588 & 0.587 & 0.559 & 0.265 & 0.347 & 0.196 \\
ToolPO & 0.457 & 0.461 & 0.468 & 0.302 & 0.524 & 0.381 & 0.528 & \textbf{0.465} & 0.459 & 0.281 \\
\rowcolor{gray!10}
ToolQP & 0.474 & 0.487 & 0.498 & 0.216 & \underline{0.599} & 0.538 & 0.580 & 0.344 & 0.287 & \textbf{0.314} \\
\rowcolor{table-blue}
\tool & \textbf{0.531} & \underline{0.527} & \textbf{0.546} & \textbf{0.316} & \underline{0.599} & \underline{0.617} & \underline{0.586} & \underline{0.422} & \underline{0.491} & \underline{0.284} \\
\bottomrule
\end{tabular}
}
\vspace{-1.0em}
\end{table}

\section{Experimental Analysis}

\subsection{Main Results}
To evaluate the effectiveness of \tool~in large-scale tool repositories, we conduct tool selection experiments on StableToolBench using Qwen2.5-7B-Instruct and Qwen3-4B-Instruct. 
Table~\ref{main_table} reports the experimental results of our model and the compared baselines.
The experimental results show that:
(1) \tool~achieves substantial overall performance gains across backbone LLMs of different scales on all evaluation metrics, indicating both its effectiveness and its feasibility for smaller-scale LLMs.
For example, on Qwen2.5-7B-Instruct, \tool~improves the overall F1 score from 9.8\% to 51.3\% and the Match score from 4.6\% to 27.8\% compared with the multi-turn baseline.
Similarly, on Qwen3-4B-Instruct, \tool~increases the overall F1 score from 40.8\% to 53.1\% and the Match score from 16.9\% to 31.6\%.
(2) Compared with other training methods, \tool~achieves substantial and stable improvements across both backbone models in overall performance, particularly in intra-collection multi-tool scenarios (I3) and unseen-instruction generalization (Inst.).
For example, on Qwen2.5-7B-Instruct, \tool~outperforms the second-best method (\textit{i.e.}, GDPO) by 1.7\% in overall F1. 
\tool~also achieves the highest F1-Inst. scores across all three scenarios, reaching 59.5\%, 48.5\%, and 29.4\% on I1, I2, and I3, respectively. 
This advantage is especially notable in the challenging I3 scenario, where \tool~surpasses GDPO, MARAG-R1, GSPO, and Search-R1 by 6.6\%, 8.3\%, 9.8\%, and 10.0\%, respectively.
A similar performance trend can be observed on Qwen3-4B-Instruct, further demonstrating the stable improvements achieved by our method.
\textbf{Overall, \tool~shows substantial and stable tool selection performance compared with other methods, especially on unseen instructions and difficult scenarios where complex tool search and compatible tool combinations are required.} 

\subsection{Tool-Selection Capabilities Across Diverse Scenarios}

\begin{figure*}[ht!]
\centering
\includegraphics[width=\textwidth]{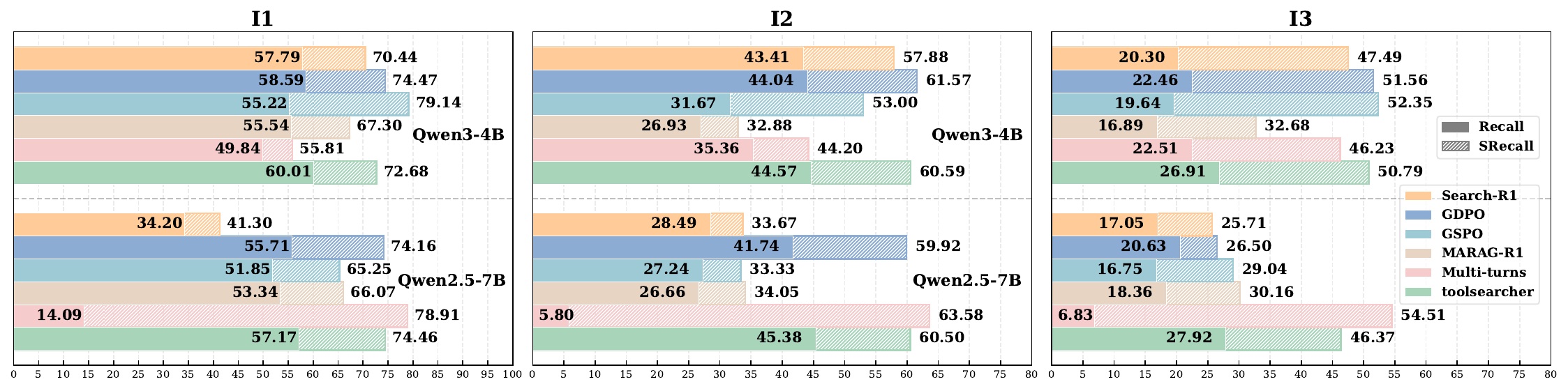}
\vspace{-2.0em}
\caption{Tool-Selection Capabilities Across Diverse Scenarios.
} 
\label{Various}
\end{figure*}

Given that tool selection involves balancing search-stage retrieval and final-stage decision-making, we analyze model performance using SRecall and Recall across the three scenarios of StableToolBench. 
As illustrated in Fig.~\ref{Various}, the experimental results indicate that \textbf{our method effectively balances these two capabilities across different backbone LLMs, thereby improving final tool-selection performance}:
(1) For Qwen-4B-Instruct, our method improves both search-stage and final-selection recall over the base model, achieving consistent gains in final tool-selection performance compared with other methods across the three scenarios. 
Notably, in scenario I3, our method improves Recall to 26.91\%, whereas other methods degrade the base model’s final-selection performance while shifting their focus to the search stage.
(2) Qwen-7B-Instruct shows high search-stage recall but weak final-selection ability, as it struggles to understand and choose among the retrieved tools. Our method slightly reduces SRecall but substantially improves the model’s ability to interpret retrieved results and make final decisions, yielding Recall improvements of 43.08\%, 39.58\%, and 21.09\% on I1/I2/I3, respectively. 
Compared with other RL methods, our method improves tool understanding and final selection, targeting the core missing ability and achieving a stronger overall trade-off.

\subsection{Generalization Capability to Out-of-Distribution}
\begin{wraptable}{r}{0.60\columnwidth}
\vspace{-1.0em}
\caption{Comparison of RL Methods Performance on the Out-of-Distribution Dataset AppWorld.}
\label{appworld}
\centering
\resizebox{0.60\textwidth}{!}{
\footnotesize
\begin{tabular}{lccccccccccc}
\toprule
\multirow{3}{*}{Method} 
  & \multicolumn{4}{c}{Tool Selcetion} 
  & \multicolumn{6}{c}{Tool Calling} \\
\cmidrule(lr){2-5} \cmidrule(lr){6-11} 
  & \multirow{2}{*}{F1} & \multirow{2}{*}{Recall} & \multirow{2}{*}{Precision} & \multirow{2}{*}{SRecall}& 
  \multicolumn{2}{c}{Avg} &\multicolumn{2}{c}{Difficulty-1} & \multicolumn{2}{c}{Difficulty-2} \\
  &&&&& TGC &SGC&TGC &SGC&TGC &SGC \\
\midrule
\rowcolor{gray!10}
\multicolumn{11}{c}{Qwen2.5-7B-Instruct}\\
Multi-turns&0.327&0.263&0.483&0.428&0.133&0.029& 0.140&0.053&0.125&0.00\\
\rowcolor{gray!10}
Search-R1 &0.455&0.381&0.640&0.591&0.181&0.086&0.298&0.158&0.042&0.00\\
GSPO  &0.478&0.411&\underline{0.657} & 0.601&0.123&0.057&0.175&0.105&0.062&0.00\\
\rowcolor{gray!10}
GDPO     &\underline{0.483}&\underline{0.425}&0.626&\underline{0.693}&\underline{0.277}&\underline{0.171}&\underline{0.351}&\underline{0.263}&\textbf{0.188}&\textbf{0.062}\\
MARAG-R1  &0.463&0.402&0.614&0.602&0.104&0.029&0.140&0.053&0.062&0.00\\
\rowcolor{gray!10}
\rowcolor{table-blue}
\tool &\textbf{0.514}&\textbf{0.444}&\textbf{0.670}&\textbf{0.697}&\textbf{0.334}&\textbf{0.228}&\textbf{0.456}&\textbf{0.368}&\textbf{0.188}&\textbf{0.062} \\
\midrule
\rowcolor{gray!10}
\multicolumn{11}{c}{Qwen3-4B-Instruct}\\
Multi-turns&0.436& 0.355&0.615&0.435 &0.152&0.086&0.228&0.158&0.062&0.00\\
\rowcolor{gray!10}
Search-R1&\underline{0.555}&\underline{0.529}&\underline{0.617}&\underline{0.635}&\underline{0.314}&\underline{0.228}&\underline{0.456}&\underline{0.368}&0.146&0.062\\
GSPO &0.400&0.317&\textbf{0.635}&0.375&0.133&0.057&0.193&0.105&0.062&0.00\\
\rowcolor{gray!10}
GDPO&0.475&0.421&0.611&0.568&0.296&0.229&0.404&0.316&\underline{0.167}&\textbf{0.125}\\
MARAG-R1 &0.395&0.314&0.598&0.369&0.134&0.057&0.211&0.105&0.042&0.00\\
\rowcolor{gray!10}
\rowcolor{table-blue}
\tool &\textbf{0.562}&\textbf{0.543}&0.612&\textbf{0.664}&\textbf{0.372}&\textbf{0.257}& \textbf{0.526}&\textbf{0.368}&\textbf{0.188}&\textbf{0.125}\\
\bottomrule
\end{tabular}
}
\end{wraptable}
We evaluate different RL methods on the out-of-distribution (OOD) benchmark AppWorld, where OOD refers to a distribution shift in task scenarios and tool usage patterns, particularly involving stateful app tools. 
As shown in Table~\ref{appworld}, our method consistently achieves the best overall performance across both model scales. For tool selection, \tool~obtains the highest F1, Recall, and SRecall on both LLMs, indicating more accurate and stable tool search and decision-making under distribution shifts. 
For tool calling performance, our method also achieves the best Task Goal Completion (TGC) and Scenario Goal Completion (SGC) on average. 
For example, \tool~outperforms the second-best method (\textit{i.e.}, GDPO) by 5.7\% in both overall TGC and SGC on Qwen2.5-7B-Instruct.
Similarly, on Qwen3-4B-Instruct,  \tool~outperforms the second-best method (\textit{i.e.}, Search-R1) by 5.8\% and 2.9\% in TGC and SGC, respectively.
It maintains consistent and better performance compared to all baseline methods across both Difficulty-1 and Difficulty-2 settings. In particular, in the highly challenging Difficulty-2 setting, which involves up to 8 unique APIs, our method consistently outperforms all baselines, while most baselines (\textit{e.g.}, Search-R1, GSPO, and MARAG-R1) achieve near-zero SGC.
Overall, compared with other methods, our method achieves more consistent performance in both tool selection and downstream execution under OOD scenarios.
Further experiments are provided in Appendix~\ref{appendix_Further}.

\subsection{Ablation Study and Method Analysis}

\begin{figure*}[ht!]
\centering
\includegraphics[width=0.80\linewidth]{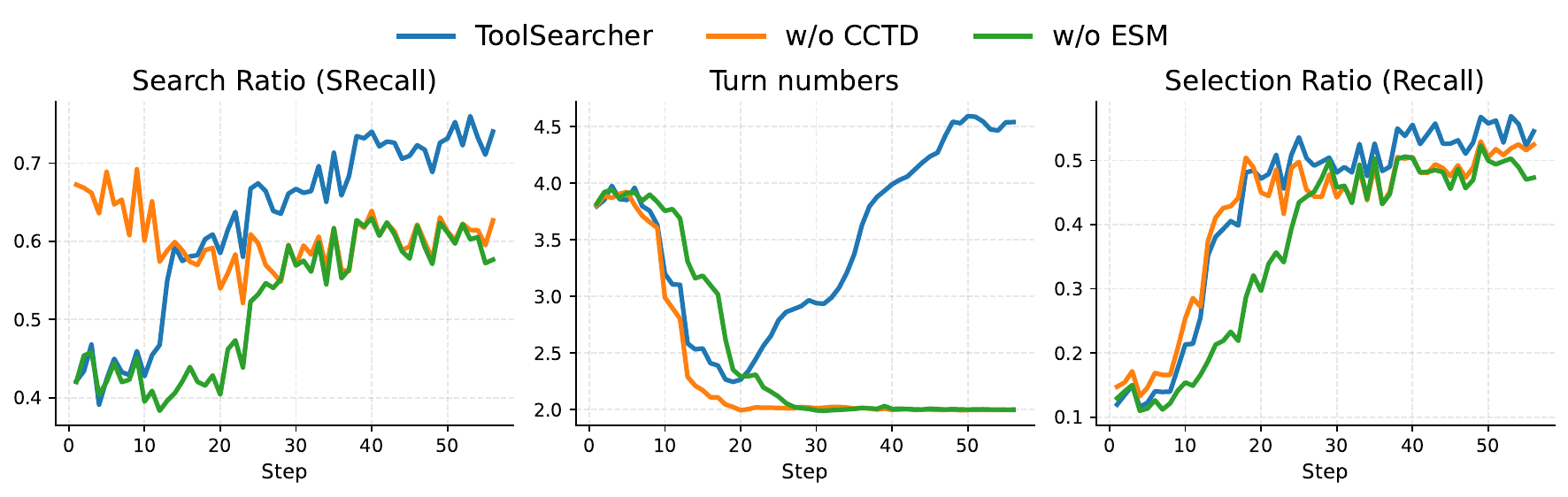}
\vspace{-1.0em}
\caption{Training Curves of Metrics.}
\vspace{-1.0em}
\label{fig:Curves}
\end{figure*}

As illustrated in Table~\ref{tab:ablation}, we perform an ablation study to evaluate the contributions of individual components of our approach, namely Category-Constrained Tool Discrimination (CCTD), Event-level Search Modeling (ESM), and Trajectory-Aligned Credit Allocation (TCA), by removing each component separately on Qwen2.5-7B-Instruct.
The experimental results show that: 
(1) No matter which component we drop, it hurts the overall performance of our model, which signals the effectiveness of all three components in tool selection. 
(2) Removing TCA disables reward assignment based on trajectory-level progress, resulting in consistent performance drops across both datasets. 
\begin{wraptable}{r}{0.57\columnwidth}
\vspace{-1.0em}
\centering
\footnotesize
\caption{Ablation Study.}
\label{tab:ablation}
\resizebox{\linewidth}{!}{
\begin{tabular}{clcccc}
\toprule
Datasets & Method & F1 & Recall & Precision & SRecall \\
\midrule
\multirow{4}{*}{StableToolBench} & \textbf{\tool} & \textbf{0.513}& \textbf{0.511} & \textbf{0.534} & \textbf{0.680} \\
& w/o CCTD & 0.477 & 0.466 & 0.510 & 0.583 \\
& w/o ESM &  0.394& 0.377&0.433&0.456 \\
& w/o TCA & 0.461 & 0.444 & 0.501 & 0.538 \\
\midrule
\multirow{4}{*}{AppWorld} & \textbf{\tool} & \textbf{0.514}& \textbf{0.444}& \textbf{0.669} & \textbf{0.697}\\
& w/o CCTD & 0.433 & 0.375 & 0.576 & 0.452 \\
& w/o ESM & 0.328 & 0.280 & 0.447 & 0.303 \\
& w/o TCA & 0.469 & 0.431 & 0.567 & 0.565 \\
\bottomrule
\end{tabular}
}

\caption{Comparsion under different Search Setting.}
\label{Comparsion_search}
\resizebox{\linewidth}{!}{
\begin{tabular}{clcccc}
\toprule
Model & Search Setting & F1 & Recall & Precision & SRecall \\
\midrule
\multirow{2}{*}{Qwen2.5-7B-Instruct} & CCTD &0.088&0.099&0.087&0.401\\
& Global search&\textbf{0.098}&\textbf{0.110}&\textbf{0.100}&\textbf{0.724}\\
\midrule
\multirow{2}{*}{Qwen3-4B-Instruct}  & CCTD & 0.265&0.280&0.272&0.344\\
& Global search& \textbf{0.408} & \textbf{0.439} & \textbf{0.410} & \textbf{0.516} \\
\bottomrule
\end{tabular}
}
\end{wraptable}
On StableToolBench, the F1 score decreases from 51.3\% to 46.1\%, and SRecall drops from 68.0\% to 53.8\%. 
On AppWorld, the F1 score decreases from 51.4\% to 46.9\%, and SRecall drops from 69.7\% to 56.5\%. 
These results highlight that trajectory-aligned credit assignment provides more accurate and stable learning signals for tool selection at various stages of progress.
(3) Removing ESM significantly decreases performance for tool searching, with the SRecall score dropping by 22.4\% on StableToolBench and 39.4\% on AppWorld, highlighting its importance in tool searching.
In Fig.~\ref{fig:Curves}, the training curve for Qwen2.5-7B-Instruct illustrates that removing ESM reduces the average number of interaction rounds with the search engine from 4 to 2. In the absence of ESM, the model is constrained to result-based rewards, which discourage its active engagement with the search process.
In contrast, our approach leverages event-level search rewards to facilitate continuous improvement, achieving superior performance in both the search phase and final selection.
(4) As shown in Table~\ref{Comparsion_search}, we evaluate the performance of the base model under different search settings. 
In the CCTD scenario, all metrics show a significant drop compared to the global search, demonstrating the effectiveness of CCTD in constructing challenging environments. 
The ablation results in Table~\ref{tab:ablation} further confirm the importance of CCTD. 
For example, the F1 score, Recall, and Precision all exhibit substantial declines when CCTD is removed, highlighting its critical role in training for tool selection.
Detailed ablation settings and gradient analysis of our method are provided in Appendix~\ref{appendix_ablation} and~\ref{Gradient}.

\section{Conclusion}
This paper presents a novel reinforcement learning framework \tool~for tool selection in large-scale repositories. 
By category-constrained tool discrimination and event-level search modeling, we significantly enhance the LLM's ability to distinguish similar tools and optimize tool compositions through multi-turn search. Additionally, the trajectory-aligned credit allocation enables fine-grained optimization during the selection process, ensuring that the rewards correspond to the advancement made relative to overall progress.
Experimental results demonstrate that \tool~outperforms several reinforcement learning baseline methods, both in large-scale tool selection scenarios StableToolBench and out-of-distribution scenarios AppWorld. These results confirm \tool's effectiveness in tasks requiring iterative search and complex tool composition, providing valuable insights for future applications of LLMs in large-scale tool selection.

\section{Acknowledgements}
This research was supported by the ``Pioneer'' and ``Leading Goose'' R\&D Program of Zhejiang (2025C02022), the National Natural Science Foundation of China (No.62507040),  the Ningbo ``Yongjiang Talent Program'' Youth Innovation Project (2024A-156-G), Ant Group, the National Natural Science Foundation of China (No.62293555).



\bibliographystyle{plain}
\bibliography{ref} %

\appendix

\section*{Appendix}


\section{Limitations}
\label{Limitations}
Although our approach demonstrates improved performance over baseline models on the StableToolBench training set and out-of-distribution settings like AppWorld, there are several limitations to consider. 
First, the training data used in our method is synthesized based on LLMs (from Stabletoobench's training data), where the majority of tool combinations are parallel rather than sequential, even when cross-category tool compositions are involved. 
This results in a weak progression in tool combination usage, which is further exacerbated in complex scenarios. In contrast, AppWorld’s APIs are mostly stateful, meaning they alter user-related information and impact database states. Due to the lack of state-awareness in the synthesized training data, our model's performance improvement in AppWorld remains limited.

Beyond large-scale tool selection, the challenge of effectively combining stateful data and APIs with tight cross-category links remains an open problem, and we consider this as future work. 
Additionally, despite achieving superior performance over baseline models in AppWorld, our approach is constrained by the context window during training, limiting the interaction rounds to 8 (equivalent to a total of 5,000 tokens). 
However, the average number of APIs involved in an AppWorld scenario exceeds 8, which presents a challenge even though our method outperforms other RL methods.
However, the average number of APIs involved in the AppWorld scenario can exceed 8. We believe a feasible strategy to address this is to summarize interaction context information appropriately, thereby shortening the context window. Future work will focus on exploring how to effectively implement this during both training and inference to handle scenarios that involve more complex tool combinations.

\section{Dataset and Metric details}
\label{Dataset_Metric}
\subsection{StableToolbench Dataset details.} 
StableToolbench consists of 765 samples, with the maximum and mean API call counts required to complete the tasks being 6 and 2.35, respectively.
StableToolbench gather 16,464 representational state transfer (REST) APIs from RapidAPI.
RapidAPI is a leading API marketplace that connects developers with thousands of
real-world APIs, streamlining the process of integrating diverse services into applications. 
Developers
All APIs in RapidAPI can be classified into 49 coarse-grained categories, such as sports, finance, and weather. 
The categories associate an API with the most relevant topic. 
Additionally, the hub also provides 500+ fine-grained categorization called \textbf{collections} (\textit{e.g.}, Chinese APIs and database APIs.) APIs in the same collection share a common characteristic and often have similar functionalities or goals.
It evaluates the generalization ability across different scenarios: (1) \textbf{Inst.}: unseen instructions for the same set of tools, (2) \textbf{Tool}: unseen tools from the same category, and (3) \textbf{Cate.}: unseen tools from a different category than those in the training data.

StableToolbench perform experiments on three scenarios: 
\textbf{single-tool instructions (I1)}, \textbf{intra-category multi-tool instructions (I2)}, and \textbf{intra-collection multi-tool instructions (I3)}. 
For I1, it conduct the evaluation for the aforementioned three levels (I1-Inst., I1-Tool, and I1-Cat.); for I2, since the training instructions already involve different tools of the same category, it only perform level 1 and level 3 for the generalization evaluation (I2-Inst. and I2-Cat.); 
similarly, it only perform level 1 generalization for I3 (I3-Inst.) 
since it already covers instructions that involve various combinations of tools from different categories (the tools in a RapidAPI collection may come from different RapidAPI categories).

For the training data, we utilize their dataset, which is automatically constructed by sampling different combinations of APIs and crafting various instructions involving them. Specifically, they focus on two key aspects in the instruction generation process:
(1) diversity: to train LLMs to handle a wide range of API usage scenarios, thereby boosting their
generalizability and robustness; and (2) multi-tool usage: to mirror real-world situations that often demand the interplay of multiple tools, improving the practical applicability and flexibility of LLMs.
We filtered out non-English data and unnatural instructions, particularly those containing API names, as they inadvertently revealed the target tool's API, leading to a subsequent decrease in the quality of the training data.
The training dataset consists of 14,418 samples, which contains 11,329/2,253/836 samples for I1/I2/I3. Overall statistics of the dataset are given in Table~\ref{Training_Data}. 
 \begin{table}[h!]
\centering
\caption{Core Information of Training Data.}
\begin{tabular}{crc}
\toprule
\textbf{Data Source} & \textbf{Number of Samples} & \textbf{API Count (Mean/Max)} \\
\midrule
I1  & 11,329 & 2.13/5 \\
I2& 2,253  & 2.30/5   \\
I3& 836  & 2.56/5 \\
Total & 14,418 & 2.18/5  \\
\bottomrule
\end{tabular}
\label{Training_Data}
\end{table}
\subsection{Appworld Dataset details.} 
AppWorld Benchmark consists of complex tasks that cover everyday scenarios using the applications in the AppWorld Engine. These tasks are natural, challenging, and diverse, and are carefully designed with distractors and hurdles to require thorough reasoning. Each task typically involves multiple applications and requires the use of many APIs in an intricate workflow, with an average of 9.5 APIs and a maximum of 26 APIs per task. Since AppWorld tasks can often be completed through multiple valid paths, the required tool sequence is not unique. 
Therefore, for tool selection evaluation, we use the AppWorld training set as test set, which provides ground-truth annotations for the involved tools, while not using it for training. 
In contrast, the original AppWorld Test-N set does not provide tool-selection labels due to the multi-solution nature of the tasks
Following the original AppWorld evaluation protocol, we evaluate tool selection on Test-N through execution outcomes, where the selected tools are used to generate and execute code, and performance is measured based on whether the final task goal is successfully completed.

AppWorld is adopted as an out-of-distribution robustness benchmark to evaluate \tool~in an unseen tool environment. 
Notably, AppWorld places greater emphasis on stateful API interactions: it simulates rich user states, and API executions may modify user data. 
For tool selection evaluation, we include 147 tasks covering 49 distinct scenarios. 
For downstream evaluation ``tool calling'', we use a subset consisting of Difficulty (D)-1 and D-2, which contains 105 tasks across 34 scenarios.
We exclude D-3 because it involves substantially more APIs, making code generation and execution by the downstream agent significantly more challenging; consequently, performance on D-3 may not accurately reflect the quality of tool selection.

\subsection{Metric details.} 
For evaluating tool selection, we assess the accuracy of the results using several metrics: \textbf{F1}, which balances precision and recall to provide a comprehensive evaluation; \textbf{Recall}, which measures the proportion of relevant tools that were selected; \textbf{Precision}, which evaluates the proportion of selected tools that are actually relevant; and \textbf{Match}, which measures the exact match between the selected tools and the ideal set. Additionally, we calculate \textbf{SRecall} to determine the recall rate during the search phase, assessing the effectiveness of finding the target tools. 

For the AppWorld evaluation, we follow their predefined method and utilize FullCodeRefl with
gpt-5-mini~\footnote{https://developers.openai.com/api/docs/models/all} as the agent to generate code and execute tasks based on the tool selection result. 
FullCodeRefl generates the entire code in one go. If execution fails, it is shown the error stack trace and is asked to reflect on its mistake (like Reflexion~\cite{shinn2023reflexion}) and retry. This repeats in a loop if it fails again.
The key metrics reported include \textbf{Task Goal Completion (TGC)}, which represents the percentage of tasks for which all evaluation tests from the scenario agent passed successfully, and \textbf{Scenario Goal Completion (SGC)}, which reflects the percentage of task scenarios for which the agent passed all evaluation tests.

\section{Baselines.}
\label{Baselines}
We choose several mainstream RL methods for comparison:

(1) \textbf{SearchR1}~\cite{jin2025search}: an RL framework that supports LLM rollouts and direct optimization with a search engine. It incorporates retrieved-token masking to stabilize RL training, multi-turn interleaved reasoning and search for complex task solving, and an effective outcome reward function based on GRPO~\cite{shao2024deepseekmath}.

(2) \textbf{GSPO}~\cite{zheng2025group}: Group Sequence Policy Optimization (GSPO), a RL algorithm for training large language models. 
The key innovation of GSPO lies in its theoretically grounded definition of importance ratio based on sequence likelihood~\cite{zheng2023click}. Additionally, GSPO computes the normalized rewards as the advantages of multiple responses to a query, ensuring the alignment between sequence-level rewarding and optimization.
In multi-turn RL scenarios, a finer-grained advantage adjustment than the sequence level may is desired. GSPO makes it possible to directly use the likelihoods returned by the inference engine for optimization, thereby avoiding the need for recomputation with the training engine. 
Following the token-level objective variant of GSPO, GSPO can be especially beneficial in scenarios like partial rollout and multi-turn RL. 

(3) \textbf{GDPO}~\cite{liu2026gdpo}: Group reward-Decoupled Normalization Policy Optimization (GDPO) is a policy optimization method designed for multi-reward RL in tool-calling scenarios. It addresses the limitation of directly applying GRPO, where different reward combinations may collapse into identical advantage values after normalization, weakening the training signal. GDPO decouples the normalization of individual rewards, better preserving their relative differences and enabling more accurate optimization over multiple tool-calling rewards, thereby improving training stability and generalization.
In our setting, we design two rewards for tool calling: the recall of the search stage and the exact match of the final results, where the latter follows the practice in Search-R1.

(4) \textbf{MARAG-R1}~\cite{luo2025marag}: MARAG-R1, a reinforcement-learned multi-tool framework that enables LLMs to dynamically coordinate multiple retrieval mechanisms for broader and more precise information access. MARAG-R1 equips the model with multiple retrieval tools and learns both how and when to use them through a two-stage training process: supervised fine-tuning followed by reinforcement learning by RLOO. This design allows the model to interleave reasoning and retrieval, progressively gathering sufficient evidence for corpus-level synthesis.
In our large-scale tool retrieval setting, we find that keyword search and the use of multiple heterogeneous retrieval tools in MARAG-R1 perform poorly, making it difficult to retrieve the target tools. For a fair comparison, we therefore use the same search engine tool as in our method. Following the original MARAG-R1 reward design, we adopt the token-level F1 score to evaluate the tool selection results, the Document Coverage Reward to assess the completeness and precision of retrieved evidence, and the Tool Exploration Reward to encourage sufficient but not excessive tool usage.

(5) \textbf{ToolPO}~\cite{li2026deepagent}: the RL-based training method introduced in DeepAgent for end-to-end interactive tool use. We adapt it to multi-turn tool search and final API selection, following its token-level policy optimization with tool-related reward signals.

(6) \textbf{ToolQP}~\cite{fang2026beyond}: a query-planning-based method that decomposes user requests into sub-tasks to guide multi-round tool retrieval. Following its original setting, we use gpt-4.1-mini to annotate training plans and incorporate plan-level similarity scores to evaluate generated plans during training.

\paragraph{Training and Inference Setup.} 
In our experiment, we carefully control the training setup to make the experimental comparison as fair as possible. 
For the RL method,  all baselines and our method use the same training configuration (e.g., same retrieval model, the number of epochs, batch size, the maximum number of documents per retrieval, the number of interaction rounds, and rollout size for each prompt). 
For supervised fine-tuning, we ensure the same overall amount of training data is observed by adjusting the number of epochs, since the RL method samples multiple answers for each prompt. 
The model is trained until convergence, and we report the best checkpoint according to validation  performance.
For the RAG paradigm, we use the top-100 retrieved documents as the context for both SFT and testing. This choice is motivated by the observation that a smaller top-k in single-round retrieval is often insufficient to recall the majority of target tools.

\section{Implementation Details}
\label{Implementation}
Following the instruction and tool-calling format of the search engine, we conduct experiments on Qwen2.5-7B-Instruct. To further examine the robustness of our method on smaller-scale models, we adopt Qwen3-4B-Instruct as an additional baseline. 
Evaluation is conducted separately on StableToolBench and AppWorld to assess in-domain and out-of-domain performance, respectively.

\paragraph{Tool Retriever for Search Engine.} We evaluated the performance of different retrievers on StableToolBench, including BM25~\cite{robertson2009probabilistic}, UniXcoder~\cite{guo2022unixcoder}, GIST-Large~\cite{solatorio2024gistembed}, Arctic-Embed 2.0~\cite{yu2024arctic}, and the Qwen-Embedding series (0.6B, 4B, 8B)~\cite{zhang2025qwen3}. Among these models, Qwen-Embedding-0.6B achieved the best performance while maintaining a relatively small parameter size, reducing the computational load during RL training. Due to its superior performance, we adopt it as the retriever to implement the search engine functionality during both training and testing.

\paragraph{Training and Inference Settings.} 
In training stage, for tool search, we use the tools from StableToolBench as the large-scale tool repository. To ensure a fair comparison while considering the training context length, we set the number of retrieved tool documents to 5 per turn for all multi-turn methods. 
Training is performed on a single node with 8 A100 (80G) GPUs. We use a total batch size of 256, and train for 1 epoch consisting of 56 steps on the training set. We train under the Category-Constrained Tool Discrimination setting for 30\% of the data (17 steps), after which we switch back to global search.
The maximum response sequence length is set to 5,000 tokens, with retrieved content capped at 768 tokens per turn.
To optimize GPU memory usage, we enable gradient checkpointing and use Fully Sharded Data Parallel (FSDP)~\cite{zhao2023pytorch} with CPU offloading.
For efficient LLM rollouts, we adopt sglang~\cite{zheng2024sglang} with a tensor parallel size of 2 and GPU memory utilization ratio of 0.5. 
The rollout sampling uses a temperature of 1.0 and a top-p value of 1.0. 
The KL divergence regularization coefficient $\beta$ and clip ratio $\epsilon$ are set to 0.001 and 0.2.
We set the policy LLM learning rate to 1e-6 and sample 5 responses per prompt, following the GRPO implementation in Verl~\cite{sheng2025hybridflow}. 
we set the maximum number of interaction turns to 8 for each prompt.
The model is trained with a learning rate warm-up ratio of 0.285. 
During inference, we keep the number of retrieved documents at 5 per turn and set the maximum response sequence length to 30,000 tokens. The maximum number of interaction turns remains the same as in training (\textit{i.e.}, 8 turns). 
To support sufficient reasoning and avoiding truncation, we increase the tokens of retrieved content to 4,096 tokens per turn.
For all RL-based training baselines, we adopt the same hyperparameter settings as described above to ensure a fair comparison.


\section{Detailed Settings of Ablation Study} 
\label{appendix_ablation}
We perform an ablation study to evaluate the contributions of individual components of our approach, namely Category-Constrained Tool Discrimination (CCTD), Event-level Search Modeling (ESM), and Trajectory-Aligned Credit Allocation (TCA), by removing each component separately.
(1) For Category-Constrained Tool Discrimination (CCTD), we remove category-level constraints and instead allow the model to perform global search over the entire tool space during the whole training process, rather than restricting search within sub-categories.
(2) For Event-level Search Modeling (ESM), we remove event-level reward modeling and replace it with trajectory-level supervision, where rewards are assigned based only on the final outcome of the entire trajectory instead of intermediate search events.
(3) For Trajectory-Aligned Credit Allocation,  we do not constrain the reward for search events that discover the same target tool at the trajectory level and remove reward correction for tool selection in trajectories where the target tool has not been fully identified.

\section{Gradient Analysis} 
\label{Gradient}

\paragraph{\tool.}
The RL objective of tool seraching using the search engine $\mathcal{R}$ is formulated  as follows:

\begin{equation}
\footnotesize
\begin{split}
    &\mathcal{J}_{\text{search}}(\theta) = \mathbb{E}{[q\sim \mathcal{D}, \{\mathcal{S}_i\}_{i=1}^G \sim \pi_{\theta_{\text{old}}}(\mathcal{S}|q,\mathcal{T}, \mathcal{R})]}  \\
    & \frac{1}{G}\sum_{i=1}^G\frac{1}{|\mathcal{S}_i|} \sum_{t=1}^{|\mathcal{S}_i|} \left\{ \min \left[ w_{i,t}(\theta,\mathcal{R}) \hat{A}_{i,e(t)}, \text{clip} \left( w_{i,t}(\theta,\mathcal{R}), 1 - \epsilon, 1 + \epsilon \right)  \hat{A}_{i,e(t)} \right] - \beta \mathbb{D}_{\text{KL}}\left(\pi_{\theta} || \pi_{\text{ref}}\right)\right\} ,
\end{split}
\end{equation}

Following the GRPO~\cite{shao2024deepseekmath}'s Gradient Analysis (assume $\pi_\theta$
 =  $\pi_{\theta_{\text{old}}}$ for simplified analysis), the gradient of $\mathcal{J}_{\text{search}}(\theta)$ is: 
\begin{equation}
\footnotesize
\begin{split}
    \nabla_{\theta}\mathcal{J}_{Search}(\theta)  & = \mathbb{E}{[q\sim \mathcal{D}, \{\mathcal{S}_i\}_{i=1}^G \sim \pi_{\theta_{\text{old}}}(\mathcal{S}|q,\mathcal{T}, \mathcal{R})]} \\
    & \frac{1}{G}\sum_{i=1}^G\frac{1}{|\mathcal{S}_i|} \sum_{t=1}^{|\mathcal{S}_i|}  
    \left[\hat{A}_{i,e(t)} + \beta \left(\frac{\pi_{ref}(\mathcal{S}_{i,t}\mid q,\mathcal{R},\mathcal{S}_{i,<t})}
{\pi_{\theta}(\mathcal{S}_{i,t}\mid q,\mathcal{R},\mathcal{S}_{i,<t})} - 1\right)\right]  \nabla_{\theta}\log \pi_\theta(\mathcal{S}_{i,t}\mid q,\mathcal{R},\mathcal{S}_{i,<t}). 
\end{split}
\end{equation}
The gradient coefficient for the token $\mathcal{S}_{i,t}$ from search-event $e(t)$ is :
\begin{equation}
\footnotesize
    GC_{search}(q, \mathcal{R}, t, \pi_{\theta}) = \hat{A}_{i,e(t)} + \beta \left(\frac{\pi_{ref}(o_{i,t}|o_{i,<t})}{\pi_{\theta}(o_{i,t}|o_{i,<t})} - 1\right),
\end{equation}
where $\hat{A}_{i,e(t)}$ is computed based on the event-level reward scores. 
The second term in the gradient expression comes from the KL divergence, which quantifies the difference between the reference policy \(\pi_{\text{ref}}\) and the current policy \(\pi_\theta\). We simplify this term to simplify the analysis. 
For different search events, the gradient contribution is computed based on the event-specific advantage \(\hat{A}_{i,e(t)}\).
From the event-level perspective, for the group of trajectory $\{\mathcal{S}_i, e_{\text{s}}^{(i)}\}_{i=1}^G$, the gradient coefficient for the search event is defined as:

\begin{equation}
\footnotesize
GC_{search}(q, \mathcal{R}, e(t), \mathcal{S}_i, \pi_{\theta}) =
\begin{cases}
\hat{A}_{i,e(t)}, & e(t)\in \mathcal{E}_i^*, \\
0, & \text{otherwise}.
\end{cases}
\end{equation}

The gradient optimization for different search events \(e(t) \in \mathcal{S}\) is controlled through the gradient coefficient $GC_{search}(q, \mathcal{R}, e(t), \mathcal{S}_i, \pi_{\theta})$.

Similarly, for selection event \(e_{\text{s}}\),  the gradient of $\mathcal{J}_{\text{select}}(\theta)$ is: 
\begin{equation}
\footnotesize
\begin{split}
\displaystyle
    &\nabla_{\theta}\mathcal{J}_{select}(\theta)   = \mathbb{E}{[e_{\text{s}}^{(i)} \sim \pi_{\theta_{\text{old}}}(e_{\text{s}}^{(i)} | q,\mathcal{S}_i)]} \\
    & \frac{1}{G}\sum_{i=1}^G\frac{1}{|e_{\text{s}}^{(i)}|} \sum_{t=1}^{|e_{\text{s}}^{(i)}|}  
    \left[\hat{A}_{i,e_{\text{s}}^{(i)}} + \beta \left(\frac{\pi_{ref}(e_{\text{s}, t}^{(i)}\mid q,\mathcal{S}_{i},e_{\text{s}, <t}^{(i)})}
{\pi_{\theta}(e_{\text{s}, t}^{(i)}\mid q,\mathcal{S}_{i},e_{\text{s}, <t}^{(i)})} - 1\right)\right]  \nabla_{\theta}\log \pi_\theta(e_{\text{s}, t}^{(i)}\mid q,\mathcal{S}_{i},e_{\text{s}, <t}^{(i)}). 
\end{split}
\end{equation}

The gradient coefficient for the selection event $e_{\text{s}}^{(i)}$ is defined as:
\begin{equation}
\footnotesize
GC_{select}(q, \mathcal{S}_i,e_{\text{s}}^{(i)}, \pi_{\theta}) =
\begin{cases}
\hat{A}_{i,e_{\text{s}}^{(i)}}, & \mathcal{M}(\mathcal{S}_i) \cap \mathcal{T}_s = \mathcal{T}_s, \\
0, & \text{otherwise}.
\end{cases}
\end{equation}
By assigning event-specific gradient coefficients and applying targeted optimization signals, the model is encouraged to learn and improve the corresponding capabilities across different stages of the overall process.

In the following, we provide a brief analysis of the gradients for the remaining methods, based on the same assumptions and simplifications.
\paragraph{GRPO.} 
The gradient coefficient for the trajectory $\{\mathcal{S}_i, e_{\text{s}}^{(i)}\}$ in GRPO is:
\begin{equation}
\footnotesize
GC_{\text{GRPO}}(q, \mathcal{R}, \mathcal{S}_i, e_{\text{s}}^{(i)}, \pi_{\theta}) = \hat{A}_{select}^{i},,
\end{equation}
where $\hat{A}_{select}$ is the normalized advantage of the final selection, computed by subtracting the mean and dividing by the standard deviation of the final selection rewards. 
The final selection reward is a binary function that equals 1 if the selection outcome matches the target tools $\mathcal{T}_s$, and 0 otherwise.
The gradient coefficient $GC_{\text{GRPO}}$ over the entire trajectory is governed by the group-wise relative advantage derived from the final selection outcome, meaning that all actions along the trajectory share the same optimization signal based on the selection result.

\paragraph{GDPO.} 
The gradient coefficient for the trajectory $\{\mathcal{S}_i, e_{\text{s}}^{(i)}\}$ in GDPO is:
\begin{equation}
\footnotesize
GC_{\text{GDPO}}(q, \mathcal{R}, \mathcal{S}_i, e_{\text{s}}^{(i)}, \pi_{\theta}) = \hat{A}_{SRecall}^{i}, + \hat{A}_{select}^{i},
\end{equation}
where $\hat{A}_{SRecall}$ is the normalized advantage computed from the search recall for target tools $\mathcal{T}_s$, reflecting the contribution of search stages to retrieving relevant tools, and $\hat{A}_{select}$ is the normalized advantage of the final selection as in GRPO.  
In GDPO, the trajectory gradient is influenced by both intermediate search performance and the final selection outcome. This means that each action along the trajectory receives a combined optimization signal. However, this also implies that gradients at different positions along the trajectory may not accurately reflect their true contribution. For example, two trajectories could have the same $\hat{A}_{SRecall}$ value but retrieve different target tools. Some search steps during the trajectory may be more critical than others, yet all steps are assigned the same reward signal, potentially leading to suboptimal credit assignment.

\paragraph{MARAG-R1.}
The gradient coefficient for the trajectory $\{\mathcal{S}_i, e_{\text{s}}^{(i)}\}$ in MARAG-R1 is:
\begin{equation}
\footnotesize
GC_{\text{MARAG-R1}}(q, \mathcal{R}, \mathcal{S}_i, e_{\text{s}}^{(i)}, \pi_{\theta}) = \hat{A}_{R(\mathcal{T})}^{i},,
\end{equation}
where $R(\mathcal{T})$ is defined as:
\begin{equation}
R(\mathcal{T}) = R_A + R_E + R_T,
\end{equation}
\textbf{Answer Reward $R_A$} evaluates the correctness of the final answer using a token-level F1 score, which provides partial credit for answers that are semantically close to the reference:
\begin{equation}
R_A = \text{F1}(A, A^*),
\end{equation}
where $A$ and $A^*$ denote the predicted and ground-truth answers, respectively.

\textbf{Document Coverage Reward $R_E$} assess the completeness and precision of retrieved evidence,  computes an F1 score over document identifiers. 
Let $\mathcal{D}_{\text{pred}}$ denote the set of retrieved document IDs and $\mathcal{D}^*$ the ground-truth supporting documents. 
The reward is defined as:
\begin{equation}
R_E = \text{F1}(\mathcal{D}_{\text{pred}}, \mathcal{D}^*) 
= \frac{2 \cdot P \cdot \text{Rec}}{P + \text{Rec}},
\end{equation}
where precision $P = \frac{|\mathcal{D}_{\text{pred}} \cap \mathcal{D}^*|}{|\mathcal{D}_{\text{pred}}|}$ 
and recall $\text{Rec} = \frac{|\mathcal{D}_{\text{pred}} \cap \mathcal{D}^*|}{|\mathcal{D}^*|}$. 
This reward encourages retrieving all necessary documents while minimizing irrelevant retrievals.

\textbf{Tool Exploration Reward $R_T$} promotes strategic exploration by rewarding sufficient but not excessive tool usage. 
Let $N_{\text{call}}$ and $N_{\text{call}}^*$ denote the number of tool calls in the predicted and expert trajectories (\textit{i.e.}, the number of rounds for invoking the search engine, the numbers of target tools), respectively:
\begin{equation}
R_T =
\begin{cases}
1, & \text{if } N_{\text{call}} \leq N_{\text{call}}^*, \\
\max\!\left(0,\, 1 - \frac{N_{\text{call}} - N_{\text{call}}^*}{N_{\text{call}}^*}\right), & \text{otherwise.}
\end{cases}
\end{equation}

These three reward components are directly summed to form the trajectory-level reward. However, simply adding them together does not highlight the model's current weaknesses in specific capabilities, especially after summation and normalization. Moreover, applying the same combined reward to all tokens along the trajectory introduces inaccurate and noisy learning signals, since different tokens and steps may contribute differently to the overall performance.

\paragraph{GSPO.}
In scenarios like multi-turn RL, GSPO use a finer-grained advantage adjustment than the sequence level to allow token-wise advantage customization:
\begin{equation}
\footnotesize
\begin{aligned}
&\mathcal{J}_\text{GSPO-token}(\theta) = \\
&\mathbb{E}_{ q \sim \mathcal{D},\, \{y_i\}_{i=1}^G \sim \pi_{\theta_\text{old}}( \cdot | q,\mathcal{T}, \mathcal{R}) }
\left[ 
\frac{1}{G} \sum_{i=1}^{G} \frac{1}{|y_i|} \sum_{t=1}^{|y_i|} 
\min \left( s_{i,t}(\theta) \widehat{A}_{i,t},  \, \mathrm{clip} \left( s_{i,t}(\theta), 1 - {\varepsilon}, 1 + {\varepsilon} \right) \widehat{A}_{i,t} \right) 
\right],
\label{equ:gspo-token}
\end{aligned}
\end{equation}
where $y_i = \{\mathcal{S}_i, e_{\text{s}}^{(i)}\}$, $s_{i,t}(\theta)$ is defined as:
\begin{align}
s_{i,t}(\theta) 
= \mathrm{sg} \left[ s_{i}(\theta) \right]  \cdot \frac{ \pi_{\theta} (y_{i,t} | q,\mathcal{T}, \mathcal{R}, y_{i,<t}) }{ \mathrm{sg} \left[ \pi_{\theta} (y_{i,t} | q,\mathcal{T}, \mathcal{R}, y_{i,<t}) \right] },
\end{align}
and $\mathrm{sg}[\cdot]$ denotes only taking the numerical value but stopping the gradient, corresponding to the \texttt{detach} operation in PyTorch. where  the importance ratio $s_{i}(\theta)$ based on sequence likelihood:
\begin{equation}
\footnotesize
\begin{aligned}
s_{i}(\theta) = \left( \frac{ \pi_{\theta} (y_i | q,\mathcal{T}, \mathcal{R}) }{ \pi_{\theta_\text{old}} (y_i | q,\mathcal{T}, \mathcal{R})} \right)^{\frac{1}{|y_i|}}
=
\exp \left( \frac{1}{|y_i|} \sum_{t=1}^{|y_i|} \log \frac{ \pi_{\theta} (y_{i,t} | q,\mathcal{T}, \mathcal{R}, y_{i,<t}) }{ \pi_{\theta_\text{old}} (y_{i,t} | q,\mathcal{T}, \mathcal{R},y_{i,<t})} \right).
\end{aligned}
\end{equation}
The gradient of GSPO can be derived as:
\begin{equation}
\scriptsize
\begin{aligned}
&\nabla_{\theta} \mathcal{J}_\text{GSPO-token}(\theta)
=\ 
\nabla_{\theta} \mathbb{E}_{ q \sim \mathcal{D},\, \{y_i\}_{i=1}^G \sim \pi_{\theta_\text{old}}( \cdot | q,\mathcal{T}, \mathcal{R}) }
\left[ 
\frac{1}{G} \sum_{i=1}^{G}
\frac{1}{|y_i|} \sum_{t=1}^{|y_i|} 
s_{i,t}(\theta) \widehat{A}_{i,t}
\right] \\
&=\ 
\mathbb{E}_{ x \sim \mathcal{D},\, \{y_i\}_{i=1}^G \sim \pi_{\theta_\text{old}}( \cdot | q,\mathcal{T}, \mathcal{R}) }
\left[ 
\frac{1}{G} \sum_{i=1}^{G} s_i (\theta)
\cdot \frac{1}{|y_i|} \sum_{t=1}^{|y_i|} 
\widehat{A}_{i,t} \frac{ \nabla_{\theta} \pi_{\theta} (y_{i,t} | q,\mathcal{T}, \mathcal{R}, y_{i,<t}) }{ \pi_{\theta} (y_{i,t} | q,\mathcal{T}, \mathcal{R}, y_{i,<t}) }
\right] \\
&=\ 
\mathbb{E}_{ x \sim \mathcal{D},\, \{y_i\}_{i=1}^G \sim \pi_{\theta_\text{old}}( \cdot | q,\mathcal{T}, \mathcal{R}) }
\left[ 
\frac{1}{G} \sum_{i=1}^{G}  \left( \frac{ \pi_{\theta} (y_i | q,\mathcal{T}, \mathcal{R}) }{ \pi_{\theta_\text{old}} (y_i | q,\mathcal{T}, \mathcal{R})} \right)^{\frac{1}{|y_i|}}
\cdot \frac{1}{|y_i|} \sum_{t=1}^{|y_i|}
\widehat{A}_{i,t} \nabla_{\theta} \log \pi_{\theta} (y_{i,t} | q,\mathcal{T}, \mathcal{R}, y_{i,<t})  
\right].
\end{aligned}
\end{equation}
Note that the term $\frac{ \pi_{\theta} (y_{i,t} | q,\mathcal{T}, \mathcal{R}, y_{i,<t}) }{ \mathrm{sg} \left[ \pi_{\theta} (y_{i,t} | q,\mathcal{T}, \mathcal{R}, y_{i,<t}) \right] }$ has a numerical value of 1, so $s_{i,t}(\theta)$ is numerically equal to $s_{i}(\theta)$.
GSPO set the advantages of all the tokens in the response $y_i$ to the same value (i.e., $\widehat{A}_{i,t} = \widehat{A}_{select}^{i},$).

The gradient coefficient for the trajectory $\{\mathcal{S}_i, e_{\text{s}}^{(i)}\}$ in GSPO is:
\begin{equation}
\footnotesize
GC_{\text{GSPO}}(q, \mathcal{R}, \mathcal{S}_i, e_{\text{s}}^{(i)}, \pi_{\theta}) = \left( \frac{ \pi_{\theta} (y_i | q,\mathcal{T}, \mathcal{R}) }{ \pi_{\theta_\text{old}} (y_i | q,\mathcal{T}, \mathcal{R})} \right)^{\frac{1}{|y_i|}} * \widehat{A}_{i,t},
\end{equation}

where GSPO adjusts the advantages for each token through the token-wise scaling factor $s_{i,t}(\theta)$. Although GSPO performs fine-grained optimization at the token level from the perspective of the entire trajectory, its optimization objective is still evaluated based on trajectory-level rewards. As a result, non-critical parts of the trajectory may still receive similar reward signals, even if they contribute little to the final outcome, which can introduce redundant or noisy optimization signals.

\section{Prompt} 
\label{Prompt}
In this section, we detail the prompts used in our experiments.
\begin{center}
\begin{tcolorbox}[title={Prompt for LLM}]
{
\textbf{System Prompt}\\
You are a super intelligent AI assistant that achieves my day-to-day tasks completely autonomously by interacting with apps/tools using their associated APIs on my behalf.\\

\textbf{Tool Schema}\\
\{tool\_schema\}  \\

\textbf{User Prompt}

Your job is to provide a list of tool API names that can be used to solve the given task, using a search engine to look up the necessary API documentation like an experienced programmer. 
You can call the search tool ``api\_doc\_search\_tool'', 
which will return the top searched results from the tools' API documentation, enclosed between < tool\_response> and </tool\_response>.
You can search as many times as you want, but include at most one search per turn. 
If you find no further external tool APIs needed, you can provide the final list of tool API names inside <tool\_list> and </tool\_list>. 
Format each tool API name as [category\_name].[tool\_name].[api\_name], and separate multiple names with commas. \\
Example: \\
<tool\_list>\\
Media\textit{.}open\_library\textit{.}osearch\_author,
Media\textit{.}movie\_db\textit{.}get\_movie\_details\\
</tool\_list>\\
Please return API names based only on the search results, and do not fabricate any.\\

\# Description of Tool Categories:\\
Here are the tool categories available for search:\\
\{tool\_categories\}  \\

\# Task Question:\\
\{question\}
}
\end{tcolorbox}
\end{center}

\begin{center}
\begin{tcolorbox}[title={Search Engine's Schema (Global Search)}]

\begin{verbatim}
tool_schema:
  type: function
  function:
    name: api_doc_search_tool
    description: Searches relevant tools' API documents
      based on the given query.
    parameters:
      type: object
      properties:
        query:
          type: string
          description: A fully-formed semantic query.
            The tool will return the top searched
            results from the tools' API documentation
            for the query.
      required:
        - query
\end{verbatim}

\end{tcolorbox}
\end{center}

\begin{center}
\begin{tcolorbox}[title={Search Engine's Schema (Category-based Search)}]
\begin{verbatim}
tool_schema:
  type: function
  function:
    name: api_doc_search_tool
    description: Searches relevant tools' API documents based on the 
    given query and optional category filter.
    parameters:
      type: object
      properties:
        category:
          type: string
          description: Optional category to filter the tool's API 
          documents. If set to "all", searches across all categories.
        query:
          type: string
          description: A fully-formed semantic query. The tool will 
          return the top searched results from the tools' API 
          documentation for the query.
      required: 
        - query
\end{verbatim}

\end{tcolorbox}
\end{center}

\section{Broader Impact}
\label{Broader_impact}
This work has the potential to positively impact the development of reliable LLM-based agents by improving their ability to search for, distinguish, and select appropriate tools from large-scale tool repositories. Since real-world tool ecosystems contain a vast number of diverse and functionally similar tools, more effective tool selection can help agents better solve complex tasks, reduce failures caused by incompatible tool composition, and make external tool use more scalable and practical. Our proposed framework may therefore benefit task that require accurate tool orchestration. At the same time, improving the tool-selection capability of LLM agents may also introduce negative societal risks. More capable agents could be misused to automate harmful actions, invoke inappropriate or unsafe tools, or interact with external environments without sufficient human oversight. In addition, errors in tool selection may still lead to unreliable outcomes, privacy leakage, or unintended execution behaviors when deployed in real-world systems. Therefore, we emphasize that such methods should be used with proper safeguards, including tool-access control, sandboxed execution, security and privacy filtering, and human supervision in high-stakes scenarios.
In conclusion, we posit that the benefits of our approach surpass its drawbacks concerning social impact.

\section{Further Experiment} 
\label{appendix_Further}

\subsection{Performance under Different Tool Library Sizes}
\label{sec:tool_library_sizes}
To evaluate the effect of tool library size, we conduct additional experiments with 4k and 8k tools using Qwen3-4B-Instruct, reporting F1, Recall, Precision, and Match.
The full 16k setting is reported in Table~\ref{main_table} and is not repeated here.
We exclude the 1k setting because the benchmark contains 1,125 unique ground-truth tools, so a library of this size would necessarily omit some target tools.

\begin{table}[ht!]
\caption{Performance under different tool library sizes on StableToolBench using Qwen3-4B-Instruct.}
\label{tab:tool_library_sizes}
\centering
\resizebox{0.90\textwidth}{!}{
\begin{tabular}{lcccccccc}
\toprule
\multirow{2}{*}{Method} & \multicolumn{4}{c}{4k} & \multicolumn{4}{c}{8k} \\
\cmidrule(lr){2-5} \cmidrule(lr){6-9}
 & F1 & Recall & Precision & Match & F1 & Recall & Precision & Match \\
\midrule
RAG & 0.496 & 0.491 & 0.517 & 0.331 & 0.415 & 0.454 & 0.413 & 0.248 \\
\rowcolor{gray!10}
RAG\textsubscript{SFT} & 0.553 & 0.565 & 0.561 & 0.340 & 0.440 & 0.493 & 0.432 & 0.275 \\
Multi-turns & 0.573 & 0.601 & 0.579 & 0.299 & 0.520 & 0.542 & 0.528 & 0.256 \\
\rowcolor{gray!10}
Search-R1 & \underline{0.696} & \textbf{0.695} & 0.711 & \underline{0.472} & 0.639 & 0.638 & 0.653 & \underline{0.420} \\
GSPO & 0.690 & 0.692 & 0.703 & 0.462 & 0.643 & \underline{0.644} & 0.655 & 0.417 \\
\rowcolor{gray!10}
GDPO & 0.692 & 0.683 & \underline{0.717} & 0.458 & \underline{0.644} & 0.637 & \underline{0.664} & 0.412 \\
MARAG-R1 & 0.632 & 0.599 & 0.705 & 0.349 & 0.577 & 0.543 & 0.649 & 0.301 \\
\rowcolor{table-blue}
\tool & \textbf{0.700} & \underline{0.693} & \textbf{0.721} & \textbf{0.478} & \textbf{0.651} & \textbf{0.646} & \textbf{0.672} & \textbf{0.429} \\
\bottomrule
\end{tabular}
}
\end{table}

As shown in Table~\ref{tab:tool_library_sizes} and Table~\ref{main_table}, the overall F1 scores of all compared methods decrease as the tool library grows from 4k to 8k and then to 16k, indicating the increased difficulty of tool selection in larger repositories.
Nevertheless, \tool~achieves the highest F1 and Match scores at all three scales and consistently improves over the Multi-turns baseline, demonstrating its effectiveness across different tool library sizes.

\subsection{Effect of Per-Round Retrieval Size}
\label{sec:per_round_retrieval_size}
We evaluate the effect of the number of tool documents retrieved per search round using Qwen3-4B-Instruct, reporting F1 and Match for $k \in \{5, 10, 20, 40\}$.
Our model is trained with $k=5$, while the per-round retrieval size is varied during evaluation.

\begin{table}[ht!]
\caption{Effect of per-round retrieval size on StableToolBench using Qwen3-4B-Instruct.}
\label{tab:per_round_retrieval_size}
\centering
\resizebox{0.90\textwidth}{!}{
\begin{tabular}{lcccccccc}
\toprule
\multirow{2}{*}{Method} & \multicolumn{2}{c}{Top-5} & \multicolumn{2}{c}{Top-10} & \multicolumn{2}{c}{Top-20} & \multicolumn{2}{c}{Top-40} \\
\cmidrule(lr){2-3} \cmidrule(lr){4-5} \cmidrule(lr){6-7} \cmidrule(lr){8-9}
 & F1 & Match & F1 & Match & F1 & Match & F1 & Match \\
\midrule
Qwen3-4B-Instruct & 0.387 & 0.220 & 0.393 & 0.167 & 0.382 & 0.171 & 0.189 & 0.065 \\
\rowcolor{gray!10}
Search-R1 & 0.505 & \underline{0.307} & 0.519 & \underline{0.327} & 0.526 & \underline{0.348} & 0.391 & 0.187 \\
GSPO & 0.473 & 0.258 & 0.493 & 0.325 & 0.524 & 0.328 & \underline{0.393} & \underline{0.208} \\
\rowcolor{gray!10}
GDPO & \underline{0.518} & 0.305 & \underline{0.525} & 0.323 & \underline{0.529} & 0.331 & 0.382 & 0.199 \\
MARAG-R1 & 0.465 & 0.225 & 0.475 & 0.244 & 0.482 & 0.242 & 0.352 & 0.195 \\
\rowcolor{table-blue}
\tool & \textbf{0.531} & \textbf{0.316} & \textbf{0.535} & \textbf{0.336} & \textbf{0.540} & \textbf{0.353} & \textbf{0.412} & \textbf{0.231} \\
\bottomrule
\end{tabular}
}
\end{table}

As shown in Table~\ref{tab:per_round_retrieval_size}, \tool~achieves the highest F1 and Match scores across all tested retrieval sizes.
Increasing $k$ from 5 to 20 improves its F1 from 0.531 to 0.540 and Match from 0.316 to 0.353, suggesting that moderately expanding the candidate set can benefit tool selection.
However, increasing $k$ to 40 substantially reduces performance across all methods.
A larger candidate set may introduce more irrelevant or distracting tools, increasing the difficulty of distinguishing and retaining relevant tool information for final selection.
Despite this decline, \tool~maintains its advantage over the compared baselines.

\subsection{Effect of Explicit Planning and Thinking}
\label{sec:planning_thinking}
We investigate whether explicit planning or thinking improves tool selection through two complementary comparisons: adding a reasoning step to the retrieval policy and enabling the native thinking mode of general-purpose models.

\paragraph{Explicit planning before retrieval.}
We compare \tool~with ToolQP~\cite{fang2026beyond}, which decomposes requests into sub-tasks to guide multi-round retrieval using teacher-annotated plans and plan-level similarity feedback during training.
We also evaluate a per-turn reasoning variant of \tool, which generates a reasoning/planning step before each retrieval turn during both training and inference.

\begin{table}[ht!]
\caption{Comparison of explicit planning and per-turn reasoning on StableToolBench. Best and second-best values are marked within each backbone.}
\label{tab:per_turn_reasoning}
\centering
\resizebox{0.90\textwidth}{!}{
\begin{tabular}{lcccccccccc}
\toprule
\multirow{2}{*}{Method} & \multicolumn{4}{c}{Overall} & \multicolumn{3}{c}{I1} & \multicolumn{2}{c}{I2} & \multicolumn{1}{c}{I3} \\
\cmidrule(lr){2-5} \cmidrule(lr){6-8} \cmidrule(lr){9-10} \cmidrule(lr){11-11}
 & F1 & Recall & Precision & Match & Inst. & Tool & Cate. & Inst. & Cate. & Inst. \\
\midrule
\rowcolor{gray!10}
\multicolumn{11}{c}{Qwen2.5-7B-Instruct}\\
ToolQP & 0.440 & 0.493 & 0.432 & 0.222 & 0.503 & 0.450 & \underline{0.504} & 0.433 & 0.362 & 0.260 \\
\rowcolor{gray!10}
\tool~(per-turn reasoning) & \underline{0.462} & \textbf{0.526} & \underline{0.446} & \underline{0.226} & \underline{0.525} & \underline{0.493} & 0.494 & \underline{0.457} & \underline{0.384} & \textbf{0.302} \\
\rowcolor{table-blue}
\tool & \textbf{0.513} & \underline{0.511} & \textbf{0.534} & \textbf{0.278} & \textbf{0.595} & \textbf{0.561} & \textbf{0.577} & \textbf{0.485} & \textbf{0.405} & \underline{0.294} \\
\midrule
\rowcolor{gray!10}
\multicolumn{11}{c}{Qwen3-4B-Instruct}\\
ToolQP & 0.474 & 0.487 & 0.498 & 0.216 & \underline{0.599} & \underline{0.538} & \underline{0.580} & 0.344 & 0.287 & \underline{0.314} \\
\rowcolor{gray!10}
\tool~(per-turn reasoning) & \underline{0.496} & \underline{0.491} & \underline{0.517} & \textbf{0.331} & \textbf{0.615} & 0.434 & 0.538 & \textbf{0.498} & \underline{0.451} & \textbf{0.327} \\
\rowcolor{table-blue}
\tool & \textbf{0.531} & \textbf{0.527} & \textbf{0.546} & \underline{0.316} & \underline{0.599} & \textbf{0.617} & \textbf{0.586} & \underline{0.422} & \textbf{0.491} & 0.284 \\
\bottomrule
\end{tabular}
}
\end{table}

As shown in Table~\ref{tab:per_turn_reasoning}, \tool~achieves higher overall F1 than both ToolQP and the per-turn reasoning variant on both backbones.
However, explicit reasoning benefits some metrics: on Qwen3-4B-Instruct, the per-turn reasoning variant improves Match from 0.316 to 0.331, despite reducing F1 from 0.531 to 0.496.
We observe that without dedicated supervision or rewards for intermediate reasoning quality, generated thoughts often paraphrase the request rather than provide useful planning.
The effectiveness of teacher-guided decomposition can also depend on the completeness and granularity of the initial plan.
These results suggest that adding explicit planning does not consistently improve tool selection, while event-level feedback can guide useful retrieval actions without requiring teacher-annotated plans.

\paragraph{Native thinking mode.}
We further compare three general-purpose models with their native thinking mode enabled and disabled.
Table~\ref{tab:native_thinking} shows that enabling thinking reduces overall F1 for all three models: from 0.443 to 0.392 for Qwen3-14B, from 0.461 to 0.268 for Qwen3.6-27B, and from 0.423 to 0.388 for Qwen3-32B.
Although some scenario-specific scores improve, native thinking does not yield better overall tool-selection performance in these experiments.
Without task-specific training or feedback, additional reasoning may introduce unnecessary semantic expansion or divert search away from the required tool functionalities.
Together, these comparisons suggest that effective reasoning should be aligned with the tool-selection objective; simply adding explicit reasoning steps or enabling native thinking does not guarantee better retrieval and selection.

\begin{table}[ht!]
\caption{Effect of native thinking mode on StableToolBench. Best and second-best values are marked within each model.}
\label{tab:native_thinking}
\centering
\resizebox{0.90\textwidth}{!}{
\begin{tabular}{llcccccccccc}
\toprule
\multirow{2}{*}{Model} & \multirow{2}{*}{Thinking} & \multicolumn{4}{c}{Overall} & \multicolumn{3}{c}{I1} & \multicolumn{2}{c}{I2} & \multicolumn{1}{c}{I3} \\
\cmidrule(lr){3-6} \cmidrule(lr){7-9} \cmidrule(lr){10-11} \cmidrule(lr){12-12}
 & & F1 & Recall & Precision & Match & Inst. & Tool & Cate. & Inst. & Cate. & Inst. \\
\midrule
Qwen3-14B & Off & \textbf{0.443} & \textbf{0.492} & \textbf{0.430} & \textbf{0.182} & \textbf{0.537} & \textbf{0.446} & \textbf{0.485} & \textbf{0.407} & \textbf{0.388} & \textbf{0.256} \\
\rowcolor{gray!10}
Qwen3-14B & On & \underline{0.392} & \underline{0.400} & \underline{0.408} & \underline{0.161} & \underline{0.473} & \underline{0.414} & \underline{0.435} & \underline{0.364} & \underline{0.335} & \underline{0.174} \\
\midrule
Qwen3.6-27B & Off & \textbf{0.461} & \textbf{0.490} & \textbf{0.466} & \textbf{0.247} & \textbf{0.537} & \textbf{0.514} & \textbf{0.529} & \textbf{0.418} & \textbf{0.374} & \textbf{0.198} \\
\rowcolor{gray!10}
Qwen3.6-27B & On & \underline{0.268} & \underline{0.275} & \underline{0.280} & \underline{0.126} & \underline{0.265} & \underline{0.307} & \underline{0.237} & \underline{0.324} & \underline{0.282} & \underline{0.129} \\
\midrule
Qwen3-32B & Off & \textbf{0.423} & \textbf{0.475} & \underline{0.410} & \textbf{0.178} & \textbf{0.496} & \textbf{0.459} & \textbf{0.476} & \underline{0.374} & \textbf{0.360} & \underline{0.213} \\
\rowcolor{gray!10}
Qwen3-32B & On & \underline{0.388} & \underline{0.400} & \textbf{0.411} & \underline{0.152} & \underline{0.471} & \underline{0.362} & \underline{0.422} & \textbf{0.388} & \underline{0.349} & \textbf{0.231} \\
\bottomrule
\end{tabular}
}
\end{table}

\subsection{Task Diversity and Multi-Aspect Tool Requirements}
\label{sec:task_diversity}
Large-scale tool selection must accommodate requests from diverse domains rather than a small, fixed set of applications.
Moreover, an individual request can require multiple tools serving different aspects of the user's goal.
Table~\ref{tab:task_examples} illustrates these properties with examples from the StableToolBench test set.
\begin{table}[ht!]
\caption{Examples of diverse, multi-aspect requests from StableToolBench (abbreviated). Tool names identify the API families required for each request.}
\label{tab:task_examples}
\centering
\footnotesize
\setlength{\tabcolsep}{4pt}
\renewcommand{\arraystretch}{1.15}
\begin{tabular}{@{}p{0.11\textwidth}p{0.40\textwidth}p{0.27\textwidth}p{0.16\textwidth}@{}}
\toprule
\textbf{Domain} & \textbf{User request} & \textbf{Required tool families} & \textbf{Distinct aspects} \\
\midrule
\textbf{Travel} & Planning a Thailand vacation with family, interested in scuba diving. Recommend dive sites in Thailand, and find top-tier stays in that area for accommodation. & \nolinkurl{world_scuba_diving_sites} (dive sites)\newline \nolinkurl{airbnb_v2} (stays) & activity venue\newline lodging \\[3pt]
\rowcolor{gray!10}
\textbf{Weather} & Planning a hiking trip to the Swiss Alps. Fetch the climate data for the location by lat/lon, and the moonrise/moonset times during our trip. & \nolinkurl{climate_data} (2 APIs)\newline \nolinkurl{stormglass_complete} (astronomy) & climate\newline astronomy \\[3pt]
\textbf{Food} & Planning a surprise dinner. Recommend low-carb chicken recipes, and give the nutritional info of milk --- calories, protein, fat per 100g. & \nolinkurl{recipe_search_and_diet} (recipes)\newline \nolinkurl{food_nutrional_data} (nutrition) & recipe\newline nutrition \\[3pt]
\rowcolor{gray!10}
\textbf{Movies / Entertainment} & Find a new TV show to watch with family. Suggest a high-IMDb family-friendly series, check its streaming availability on Netflix/Disney+/Apple TV, and provide cast \& crew info. & \nolinkurl{ott_details} (show details)\newline \nolinkurl{streaming_availability} (streaming)\newline \nolinkurl{watchmode} (cast \& crew) & details\newline streaming\newline cast \\[3pt]
\bottomrule
\end{tabular}
\end{table}
\begin{table}[th]
\caption{Multi-turn search with direction adjustment for the travel request. Each search returns five API candidates; the final row shows the selected APIs.}
\label{tab:multi_turn_case_study}
\centering
\footnotesize
\setlength{\tabcolsep}{4pt}
\renewcommand{\arraystretch}{1.15}
\begin{tabular}{@{}p{0.05\textwidth}p{0.40\textwidth}p{0.51\textwidth}@{}}
\toprule
\textbf{Step} & \textbf{Search query and observation} & \textbf{Retrieved / selected APIs} \\
\midrule
1 & \textbf{Aspect A: dive sites.}\newline API to find scuba diving sites in Thailand\newline\emph{Observation:} The dive-site target API is retrieved; the lodging target API is still missing. & 1. \nolinkurl{world_scuba_diving_sites_api.query_divesites_by_a_country_or_a_region}\newline 2. \nolinkurl{world_scuba_diving_sites_api.query_divesites_by_gps_boundaries_for_use_with_maps}\newline 3. \nolinkurl{provinces_of_thailand.subdistrict}\newline 4. \nolinkurl{oil_thai_price.get_image_result_today_price}\newline 5. \nolinkurl{oil_thai_price.get_latest_price} \\[4pt]
\rowcolor{table-blue}
2 & \textbf{Switch to Aspect B: lodging.}\newline API to find top-tier accommodation options in Thailand for scuba diving sites\newline\emph{Observation:} The query shifts toward lodging and retrieves its target API. Both target APIs have now been found. & 1. \nolinkurl{world_scuba_diving_sites_api.query_divesites_by_a_country_or_a_region}\newline 2. \nolinkurl{world_scuba_diving_sites_api.query_divesites_by_gps_boundaries_for_use_with_maps}\newline 3. \nolinkurl{airbnb_v2.get_top_tier_stays_filter}\newline 4. \nolinkurl{airbnb_search.properties_search}\newline 5. \nolinkurl{airbnb_search.get_top_tier_stays_filter} \\[4pt]
3 & \textbf{Confirm lodging-side API family.}\newline how to combine scuba diving site search with accommodation API for Thailand\newline\emph{Observation:} The query explores a combined requirement but retrieves no additional target API. & 1. \nolinkurl{world_scuba_diving_sites_api.query_divesites_by_a_country_or_a_region}\newline 2. \nolinkurl{world_scuba_diving_sites_api.query_divesites_by_gps_boundaries_for_use_with_maps}\newline 3. \nolinkurl{airbnb_v2.get_accessibility_filters}\newline 4. \nolinkurl{airbnb_v2.test_api}\newline 5. \nolinkurl{airbnb_v2.get_amenities_filters} \\[4pt]
\rowcolor{table-blue}
4 & \textbf{Final selection.}\newline The final selection combines the two target APIs found in earlier rounds. & 1. \nolinkurl{world_scuba_diving_sites_api.query_divesites_by_a_country_or_a_region}\newline 2. \nolinkurl{airbnb_v2.get_top_tier_stays_filter} \\[4pt]
\bottomrule
\end{tabular}
\end{table}
The examples span travel, weather, food, and entertainment, illustrating the breadth of functionality a general-purpose tool repository must cover.
Within a single request, complementary requirements can also map to different tool families: finding diving sites and accommodation, for example, requires both activity-related and lodging-related tools.
A single query emphasizing one aspect may fail to retrieve tools needed for another.
These properties motivate both a broad tool repository and multi-turn search that can adjust its direction to cover the different requirements.

\subsection{Case Study of Multi-Turn Search}
\label{sec:multi_turn_case_study}
We examine the travel request in Table~\ref{tab:task_examples}: ``Planning a Thailand vacation with family, interested in scuba diving. Recommend dive sites in Thailand, and find top-tier stays in that area for accommodation.''
The two target APIs serve distinct purposes: \nolinkurl{world_scuba_diving_sites_api.query_divesites_by_a_country_or_a_region} finds diving sites by country or region, while \nolinkurl{airbnb_v2.get_top_tier_stays_filter} provides filters for top-tier accommodation.
Table~\ref{tab:multi_turn_case_study} presents the trajectory produced by \tool, comprising three search rounds followed by final tool selection.
The first query focuses on diving sites and retrieves the corresponding target API, but does not retrieve the lodging target.
The second query shifts toward accommodation and discovers the remaining target API, covering both aspects of the request.
The third search does not discover a new target API; the final selection retains the two relevant APIs from the accumulated search results.
This case illustrates how changing search direction across turns can recover a complementary tool that was absent from the initial results.

\newpage
\section*{NeurIPS Paper Checklist}

\begin{enumerate}

\item {\bf Claims}
    \item[] Question: Do the main claims made in the abstract and introduction accurately reflect the paper's contributions and scope?
    \item[] Answer: \answerYes{} 
    \item[] Justification: The main claims made in the abstract and introduction accurately reflect the
paper’s contribution, which can be found in Section~\ref{Introduction}.
    \item[] Guidelines:
    \begin{itemize}
        \item The answer \answerNA{} means that the abstract and introduction do not include the claims made in the paper.
        \item The abstract and/or introduction should clearly state the claims made, including the contributions made in the paper and important assumptions and limitations. A \answerNo{} or \answerNA{} answer to this question will not be perceived well by the reviewers. 
        \item The claims made should match theoretical and experimental results, and reflect how much the results can be expected to generalize to other settings. 
        \item It is fine to include aspirational goals as motivation as long as it is clear that these goals are not attained by the paper. 
    \end{itemize}

\item {\bf Limitations}
    \item[] Question: Does the paper discuss the limitations of the work performed by the authors?
    \item[] Answer: \answerYes{} 
    \item[] Justification: We provide it in Appendix~\ref{Limitations}.
    \item[] Guidelines:
    \begin{itemize}
        \item The answer \answerNA{} means that the paper has no limitation while the answer \answerNo{} means that the paper has limitations, but those are not discussed in the paper. 
        \item The authors are encouraged to create a separate ``Limitations'' section in their paper.
        \item The paper should point out any strong assumptions and how robust the results are to violations of these assumptions (e.g., independence assumptions, noiseless settings, model well-specification, asymptotic approximations only holding locally). The authors should reflect on how these assumptions might be violated in practice and what the implications would be.
        \item The authors should reflect on the scope of the claims made, e.g., if the approach was only tested on a few datasets or with a few runs. In general, empirical results often depend on implicit assumptions, which should be articulated.
        \item The authors should reflect on the factors that influence the performance of the approach. For example, a facial recognition algorithm may perform poorly when image resolution is low or images are taken in low lighting. Or a speech-to-text system might not be used reliably to provide closed captions for online lectures because it fails to handle technical jargon.
        \item The authors should discuss the computational efficiency of the proposed algorithms and how they scale with dataset size.
        \item If applicable, the authors should discuss possible limitations of their approach to address problems of privacy and fairness.
        \item While the authors might fear that complete honesty about limitations might be used by reviewers as grounds for rejection, a worse outcome might be that reviewers discover limitations that aren't acknowledged in the paper. The authors should use their best judgment and recognize that individual actions in favor of transparency play an important role in developing norms that preserve the integrity of the community. Reviewers will be specifically instructed to not penalize honesty concerning limitations.
    \end{itemize}

\item {\bf Theory assumptions and proofs}
    \item[] Question: For each theoretical result, does the paper provide the full set of assumptions and a complete (and correct) proof?
    \item[] Answer: \answerYes{} 
    \item[] Justification: We provide the full set of assumptions in Section~\ref{Method} and gradient analysis in Appendix~\ref{Gradient}.
    \item[] Guidelines:
    \begin{itemize}
        \item The answer \answerNA{} means that the paper does not include theoretical results. 
        \item All the theorems, formulas, and proofs in the paper should be numbered and cross-referenced.
        \item All assumptions should be clearly stated or referenced in the statement of any theorems.
        \item The proofs can either appear in the main paper or the supplemental material, but if they appear in the supplemental material, the authors are encouraged to provide a short proof sketch to provide intuition. 
        \item Inversely, any informal proof provided in the core of the paper should be complemented by formal proofs provided in appendix or supplemental material.
        \item Theorems and Lemmas that the proof relies upon should be properly referenced. 
    \end{itemize}

    \item {\bf Experimental result reproducibility}
    \item[] Question: Does the paper fully disclose all the information needed to reproduce the main experimental results of the paper to the extent that it affects the main claims and/or conclusions of the paper (regardless of whether the code and data are provided or not)?
    \item[] Answer: \answerYes{} 
    \item[] Justification: We elaborate it in Section~\ref{Experimental_Setup} and Appendix~\ref{Dataset_Metric} and ~\ref{Implementation}.
    \item[] Guidelines:
    \begin{itemize}
        \item The answer \answerNA{} means that the paper does not include experiments.
        \item If the paper includes experiments, a \answerNo{} answer to this question will not be perceived well by the reviewers: Making the paper reproducible is important, regardless of whether the code and data are provided or not.
        \item If the contribution is a dataset and\slash or model, the authors should describe the steps taken to make their results reproducible or verifiable. 
        \item Depending on the contribution, reproducibility can be accomplished in various ways. For example, if the contribution is a novel architecture, describing the architecture fully might suffice, or if the contribution is a specific model and empirical evaluation, it may be necessary to either make it possible for others to replicate the model with the same dataset, or provide access to the model. In general. releasing code and data is often one good way to accomplish this, but reproducibility can also be provided via detailed instructions for how to replicate the results, access to a hosted model (e.g., in the case of a large language model), releasing of a model checkpoint, or other means that are appropriate to the research performed.
        \item While NeurIPS does not require releasing code, the conference does require all submissions to provide some reasonable avenue for reproducibility, which may depend on the nature of the contribution. For example
        \begin{enumerate}
            \item If the contribution is primarily a new algorithm, the paper should make it clear how to reproduce that algorithm.
            \item If the contribution is primarily a new model architecture, the paper should describe the architecture clearly and fully.
            \item If the contribution is a new model (e.g., a large language model), then there should either be a way to access this model for reproducing the results or a way to reproduce the model (e.g., with an open-source dataset or instructions for how to construct the dataset).
            \item We recognize that reproducibility may be tricky in some cases, in which case authors are welcome to describe the particular way they provide for reproducibility. In the case of closed-source models, it may be that access to the model is limited in some way (e.g., to registered users), but it should be possible for other researchers to have some path to reproducing or verifying the results.
        \end{enumerate}
    \end{itemize}

\item {\bf Open access to data and code}
    \item[] Question: Does the paper provide open access to the data and code, with sufficient instructions to faithfully reproduce the main experimental results, as described in supplemental material?
    \item[] Answer: \answerYes{} 
    \item[] Justification: The datasets we use are publicly available, and the link to the released code is presented in the abstract.
    \item[] Guidelines:
    \begin{itemize}
        \item The answer \answerNA{} means that paper does not include experiments requiring code.
        \item Please see the NeurIPS code and data submission guidelines (\url{https://neurips.cc/public/guides/CodeSubmissionPolicy}) for more details.
        \item While we encourage the release of code and data, we understand that this might not be possible, so \answerNo{} is an acceptable answer. Papers cannot be rejected simply for not including code, unless this is central to the contribution (e.g., for a new open-source benchmark).
        \item The instructions should contain the exact command and environment needed to run to reproduce the results. See the NeurIPS code and data submission guidelines (\url{https://neurips.cc/public/guides/CodeSubmissionPolicy}) for more details.
        \item The authors should provide instructions on data access and preparation, including how to access the raw data, preprocessed data, intermediate data, and generated data, etc.
        \item The authors should provide scripts to reproduce all experimental results for the new proposed method and baselines. If only a subset of experiments are reproducible, they should state which ones are omitted from the script and why.
        \item At submission time, to preserve anonymity, the authors should release anonymized versions (if applicable).
        \item Providing as much information as possible in supplemental material (appended to the paper) is recommended, but including URLs to data and code is permitted.
    \end{itemize}

\item {\bf Experimental setting/details}
    \item[] Question: Does the paper specify all the training and test details (e.g., data splits, hyperparameters, how they were chosen, type of optimizer) necessary to understand the results?
    \item[] Answer: \answerYes{} 
    \item[] Justification: We elaborate it in Section~\ref{Experimental_Setup} and Appendix~\ref{Dataset_Metric} and ~\ref{Implementation}.
    \item[] Guidelines:
    \begin{itemize}
        \item The answer \answerNA{} means that the paper does not include experiments.
        \item The experimental setting should be presented in the core of the paper to a level of detail that is necessary to appreciate the results and make sense of them.
        \item The full details can be provided either with the code, in appendix, or as supplemental material.
    \end{itemize}

\item {\bf Experiment statistical significance}
    \item[] Question: Does the paper report error bars suitably and correctly defined or other appropriate information about the statistical significance of the experiments?
    \item[] Answer: \answerNo{} 
    \item[] Justification: Error bars are not reported because it would be too computationally expensive. We fix the seed for all our experiments, which ensures the reliability of our
experiments. 
    \item[] Guidelines:
    \begin{itemize}
        \item The answer \answerNA{} means that the paper does not include experiments.
        \item The authors should answer \answerYes{} if the results are accompanied by error bars, confidence intervals, or statistical significance tests, at least for the experiments that support the main claims of the paper.
        \item The factors of variability that the error bars are capturing should be clearly stated (for example, train/test split, initialization, random drawing of some parameter, or overall run with given experimental conditions).
        \item The method for calculating the error bars should be explained (closed form formula, call to a library function, bootstrap, etc.)
        \item The assumptions made should be given (e.g., Normally distributed errors).
        \item It should be clear whether the error bar is the standard deviation or the standard error of the mean.
        \item It is OK to report 1-sigma error bars, but one should state it. The authors should preferably report a 2-sigma error bar than state that they have a 96\% CI, if the hypothesis of Normality of errors is not verified.
        \item For asymmetric distributions, the authors should be careful not to show in tables or figures symmetric error bars that would yield results that are out of range (e.g., negative error rates).
        \item If error bars are reported in tables or plots, the authors should explain in the text how they were calculated and reference the corresponding figures or tables in the text.
    \end{itemize}

\item {\bf Experiments compute resources}
    \item[] Question: For each experiment, does the paper provide sufficient information on the computer resources (type of compute workers, memory, time of execution) needed to reproduce the experiments?
    \item[] Answer: \answerYes{} 
    \item[] Justification: We provide our training GPU device and training setup in Appendix~\ref{Dataset_Metric} and ~\ref{Implementation}.
    \item[] Guidelines:
    \begin{itemize}
        \item The answer \answerNA{} means that the paper does not include experiments.
        \item The paper should indicate the type of compute workers CPU or GPU, internal cluster, or cloud provider, including relevant memory and storage.
        \item The paper should provide the amount of compute required for each of the individual experimental runs as well as estimate the total compute. 
        \item The paper should disclose whether the full research project required more compute than the experiments reported in the paper (e.g., preliminary or failed experiments that didn't make it into the paper). 
    \end{itemize}
    
\item {\bf Code of ethics}
    \item[] Question: Does the research conducted in the paper conform, in every respect, with the NeurIPS Code of Ethics \url{https://neurips.cc/public/EthicsGuidelines}?
    \item[] Answer: \answerYes{}
    \item[] Justification: We have made sure to preserve the paper’s anonymity and its conformation
with NeurIPS Code of Ethics.
    \item[] Guidelines:
    \begin{itemize}
        \item The answer \answerNA{} means that the authors have not reviewed the NeurIPS Code of Ethics.
        \item If the authors answer \answerNo, they should explain the special circumstances that require a deviation from the Code of Ethics.
        \item The authors should make sure to preserve anonymity (e.g., if there is a special consideration due to laws or regulations in their jurisdiction).
    \end{itemize}

\item {\bf Broader impacts}
    \item[] Question: Does the paper discuss both potential positive societal impacts and negative societal impacts of the work performed?
    \item[] Answer: \answerYes{} 
    \item[] Justification: We discuss that in Appendix~\ref{Broader_impact}.
    \item[] Guidelines:
    \begin{itemize}
        \item The answer \answerNA{} means that there is no societal impact of the work performed.
        \item If the authors answer \answerNA{} or \answerNo, they should explain why their work has no societal impact or why the paper does not address societal impact.
        \item Examples of negative societal impacts include potential malicious or unintended uses (e.g., disinformation, generating fake profiles, surveillance), fairness considerations (e.g., deployment of technologies that could make decisions that unfairly impact specific groups), privacy considerations, and security considerations.
        \item The conference expects that many papers will be foundational research and not tied to particular applications, let alone deployments. However, if there is a direct path to any negative applications, the authors should point it out. For example, it is legitimate to point out that an improvement in the quality of generative models could be used to generate Deepfakes for disinformation. On the other hand, it is not needed to point out that a generic algorithm for optimizing neural networks could enable people to train models that generate Deepfakes faster.
        \item The authors should consider possible harms that could arise when the technology is being used as intended and functioning correctly, harms that could arise when the technology is being used as intended but gives incorrect results, and harms following from (intentional or unintentional) misuse of the technology.
        \item If there are negative societal impacts, the authors could also discuss possible mitigation strategies (e.g., gated release of models, providing defenses in addition to attacks, mechanisms for monitoring misuse, mechanisms to monitor how a system learns from feedback over time, improving the efficiency and accessibility of ML).
    \end{itemize}
    
\item {\bf Safeguards}
    \item[] Question: Does the paper describe safeguards that have been put in place for responsible release of data or models that have a high risk for misuse (e.g., pre-trained language models, image generators, or scraped datasets)?
    \item[] Answer: \answerNA{} 
    \item[] Justification: The RL training on tool selection does not impose such risk.
    \item[] Guidelines:
    \begin{itemize}
        \item The answer \answerNA{} means that the paper poses no such risks.
        \item Released models that have a high risk for misuse or dual-use should be released with necessary safeguards to allow for controlled use of the model, for example by requiring that users adhere to usage guidelines or restrictions to access the model or implementing safety filters. 
        \item Datasets that have been scraped from the Internet could pose safety risks. The authors should describe how they avoided releasing unsafe images.
        \item We recognize that providing effective safeguards is challenging, and many papers do not require this, but we encourage authors to take this into account and make a best faith effort.
    \end{itemize}

\item {\bf Licenses for existing assets}
    \item[] Question: Are the creators or original owners of assets (e.g., code, data, models), used in the paper, properly credited and are the license and terms of use explicitly mentioned and properly respected?
    \item[] Answer: \answerYes{} 
    \item[] Justification: We use only publicly available datasets and models, and we tag and cite all corresponding sources in the paper.
    \item[] Guidelines:
    \begin{itemize}
        \item The answer \answerNA{} means that the paper does not use existing assets.
        \item The authors should cite the original paper that produced the code package or dataset.
        \item The authors should state which version of the asset is used and, if possible, include a URL.
        \item The name of the license (e.g., CC-BY 4.0) should be included for each asset.
        \item For scraped data from a particular source (e.g., website), the copyright and terms of service of that source should be provided.
        \item If assets are released, the license, copyright information, and terms of use in the package should be provided. For popular datasets, \url{paperswithcode.com/datasets} has curated licenses for some datasets. Their licensing guide can help determine the license of a dataset.
        \item For existing datasets that are re-packaged, both the original license and the license of the derived asset (if it has changed) should be provided.
        \item If this information is not available online, the authors are encouraged to reach out to the asset's creators.
    \end{itemize}

\item {\bf New assets}
    \item[] Question: Are new assets introduced in the paper well documented and is the documentation provided alongside the assets?
    \item[] Answer: \answerNA{} 
    \item[] Justification: The paper does not release new assets.
    \item[] Guidelines:
    \begin{itemize}
        \item The answer \answerNA{} means that the paper does not release new assets.
        \item Researchers should communicate the details of the dataset\slash code\slash model as part of their submissions via structured templates. This includes details about training, license, limitations, etc. 
        \item The paper should discuss whether and how consent was obtained from people whose asset is used.
        \item At submission time, remember to anonymize your assets (if applicable). You can either create an anonymized URL or include an anonymized zip file.
    \end{itemize}

\item {\bf Crowdsourcing and research with human subjects}
    \item[] Question: For crowdsourcing experiments and research with human subjects, does the paper include the full text of instructions given to participants and screenshots, if applicable, as well as details about compensation (if any)? 
    \item[] Answer: \answerNA{} 
    \item[] Justification: The paper does not involve crowdsourcing nor research with human subjects.
    \item[] Guidelines:
    \begin{itemize}
        \item The answer \answerNA{} means that the paper does not involve crowdsourcing nor research with human subjects.
        \item Including this information in the supplemental material is fine, but if the main contribution of the paper involves human subjects, then as much detail as possible should be included in the main paper. 
        \item According to the NeurIPS Code of Ethics, workers involved in data collection, curation, or other labor should be paid at least the minimum wage in the country of the data collector. 
    \end{itemize}

\item {\bf Institutional review board (IRB) approvals or equivalent for research with human subjects}
    \item[] Question: Does the paper describe potential risks incurred by study participants, whether such risks were disclosed to the subjects, and whether Institutional Review Board (IRB) approvals (or an equivalent approval/review based on the requirements of your country or institution) were obtained?
    \item[] Answer: \answerNA{}
    \item[] Justification: The paper does not involve crowdsourcing nor research with human subjects.
    \item[] Guidelines:
    \begin{itemize}
        \item The answer \answerNA{} means that the paper does not involve crowdsourcing nor research with human subjects.
        \item Depending on the country in which research is conducted, IRB approval (or equivalent) may be required for any human subjects research. If you obtained IRB approval, you should clearly state this in the paper. 
        \item We recognize that the procedures for this may vary significantly between institutions and locations, and we expect authors to adhere to the NeurIPS Code of Ethics and the guidelines for their institution. 
        \item For initial submissions, do not include any information that would break anonymity (if applicable), such as the institution conducting the review.
    \end{itemize}

\item {\bf Declaration of LLM usage}
    \item[] Question: Does the paper describe the usage of LLMs if it is an important, original, or non-standard component of the core methods in this research? Note that if the LLM is used only for writing, editing, or formatting purposes and does \emph{not} impact the core methodology, scientific rigor, or originality of the research, declaration is not required.
    \item[] Answer:  \answerYes{}
    \item[] Justification: LLMs are an important component of our core method. We use an LLM as the policy model for multi-turn tool search and tool selection, and optimize it with reinforcement learning. The paper describes how the LLM is used in the proposed framework, as well as the corresponding training and inference settings. 

    \item[] Guidelines:
    \begin{itemize}
        \item The answer \answerNA{} means that the core method development in this research does not involve LLMs as any important, original, or non-standard components.
        \item Please refer to our LLM policy in the NeurIPS handbook for what should or should not be described.
    \end{itemize}

\end{enumerate}

\end{document}